\documentclass[letterpaper]{article} 
\usepackage{aaai2027}  
\nocopyright
\usepackage[hyphens]{url}  
\usepackage{graphicx} 
\usepackage{natbib}  
\usepackage{caption} 
\usepackage{algorithm}
\usepackage{algorithmicx}
\usepackage{algpseudocode}
\usepackage{xcolor}
\usepackage[table]{xcolor}
\usepackage{array}
\usepackage{amssymb}
\usepackage{multirow}
\usepackage{amsmath}
\usepackage{tikz}
\usepackage{cleveref}

\graphicspath{{figs/}}
\definecolor{oursgray}{gray}{0.92}

\newcommand{\ms}[2]{\ensuremath{#1_{\scriptscriptstyle \pm #2}}}
\newcommand{\bms}[2]{\ensuremath{\mathbf{#1}_{\scriptstyle \pm #2}}}

\usepackage{newfloat}
\usepackage{listings}
\DeclareCaptionStyle{ruled}{labelfont=normalfont,labelsep=colon,strut=off} 
\floatstyle{ruled}
\newfloat{listing}{tb}{lst}{}
\floatname{listing}{Listing}

\usepackage{booktabs}

\title{Manifold-Constrained Noise Optimization for Diverse Diffusion Sampling}

\author {
    Qitan Shi,
    Cheng Jin,
    Ziyuan Liu,
    Yuantao Gu\corresponding
}
\affiliations {
    Tsinghua University\\
    \{sqt24,jinc21,liuziyua22\}@mails.tsinghua.edu.cn, 
    gyt@tsinghua.edu.cn
}

\begin{document}

\maketitle

\begin{abstract}
Few-step distilled diffusion models generate high-quality images quickly, but often lose per-prompt diversity, producing near-identical samples across random seeds. Optimizing the initial noise at inference time offers an appealing way to recover this diversity, yet existing methods directly update the initial noise in an unconstrained Euclidean space, ignoring both the geometry of the Gaussian prior and the model's sensitivity to noise frequencies. 
They therefore introduce auxiliary quality-control objectives to maintain generation fidelity, adding compute and weighting hyperparameters while still requiring conservative updates to prevent degradation.
In this work, we propose MoNO, a training-free method that performs Manifold-constrained Noise Optimization on a low-dimensional, quality-stabilizing noise manifold. MoNO sequentially optimizes each new initial noise so that its predicted visual feature complements previous generations, while Riemannian updates on an affine low-frequency sphere preserve prior likelihood and fix unstable high-frequency components by construction. This enables large geodesic steps, removes the need for auxiliary quality-control objectives, and converges in far fewer iterations than prior noise-optimization methods. Experiments with multiple distilled text-to-image diffusion models show that MoNO consistently improves per-prompt diversity while maintaining image quality.
\end{abstract}


\section{Introduction}

\begin{figure}[t]
  \centering
  \setlength{\tabcolsep}{1pt}
  \renewcommand{\arraystretch}{0.9}
  \newcommand{\imgw}{0.225\linewidth}
  \newcommand{\sampleimg}[1]{\includegraphics[width=\imgw]{#1}}
  \begin{tabular}{@{}cccc@{}}
    \sampleimg{introduction/iid/1.jpg}
    & \sampleimg{introduction/iid/2.jpg}
    & \sampleimg{introduction/iid/3.jpg}
    & \sampleimg{introduction/iid/4.jpg} \\[-2pt]
    \sampleimg{introduction/iid/5.jpg}
    & \sampleimg{introduction/iid/6.jpg}
    & \sampleimg{introduction/iid/7.jpg}
    & \sampleimg{introduction/iid/8.jpg} \\
    \multicolumn{4}{c}{(a) sampled from i.i.d. initial noise } \\[4pt]
    \sampleimg{introduction/ours/1.jpg}
    & \sampleimg{introduction/ours/2.jpg}
    & \sampleimg{introduction/ours/3.jpg}
    & \sampleimg{introduction/ours/4.jpg} \\[-2pt]
    \sampleimg{introduction/ours/5.jpg}
    & \sampleimg{introduction/ours/6.jpg}
    & \sampleimg{introduction/ours/7.jpg}
    & \sampleimg{introduction/ours/8.jpg} \\
    \multicolumn{4}{c}{(b) sampled from optimized initial noise}
  \end{tabular}
  \caption{Generations for the prompt \textit{``cat with sunglasses''} with FLUX.2~[klein] 4B. (a) Under i.i.d.\ sampling, few-step distilled models collapse in diversity and produce visually similar images. (b) Our method changes only the initial noise, yielding more diverse generations.}
  \label{fig:introduction}
\end{figure}

Few-step distilled diffusion models generate high-quality text-to-image samples in only one or a few network evaluations, making generation fast enough to be interactive and increasingly popular in practice~\citep{sauer2024adversarial,yin2024improved,flux-2-2025}. However, they exhibit severe mode collapse, where sampling the same prompt with different random seeds still yields nearly identical images~\citep{gandikota2026distilling,wu2026diversity,li20261}, as shown in \Cref{fig:introduction}. 
This limits many applications, since repeatedly sampling a prompt is often intended to surface different possibilities. Under mode collapse, the model instead trivially repeats near-identical images and fails to explore the broader distribution of plausible outputs.
This motivates a line of inference-time methods that, given a trained diffusion model, improve the diversity of the small set of images generated per prompt at sampling time, without any retraining.

Existing inference-time methods for improving diversity can be grouped into three families. 
Guidance methods modify the denoising dynamics at inference time, for example by perturbing the conditioning signal, introducing interactions among concurrently generated samples, or pushing visual features away from reference samples~\citep{sadat2024cads,corso2024particle,vinograd2026diverse,singh2024negative}.
Search methods, by contrast, leave the sampler itself unchanged. They sample a large candidate pool and solve a group-selection problem that trades off individual quality against pairwise diversity~\citep{parmar2025scaling}.
Noise optimization methods instead act directly on the initial noise, optimizing it so that the resulting generations become more diverse~\citep{kim2025diverse,harrington2026s}.
Among these methods, noise optimization is a natural fit for few-step diffusion models. 
By acting on the initial noise, it avoids perturbing the short pretrained sampling trajectory and does not require a large candidate pool that would offset the efficiency gains of distillation. 
It also remains applicable when there are too few denoising steps for step-wise methods to act effectively.

However, existing noise-optimization methods still rely on auxiliary quality-control objectives to prevent optimized noises from producing low-quality generations. These objectives include external reward models that provide quality signals, as well as soft regularization or anchoring terms that discourage overly aggressive noise updates. They introduce additional compute and weighting hyperparameters, yet still require conservative updates to prevent quality degradation.

Our key insight is that this conservatism is not inherent to noise optimization, but rather a consequence of optimizing the initial noise in an unconstrained Euclidean space.
The constraints that previous methods impose softly have natural hard-constrained counterparts. 
First, for an isotropic Gaussian prior, the probability density depends only on the noise norm, and high-dimensional samples concentrate in a thin annulus around their typical radius~\citep{vershynin2020high}. Thus, instead of penalizing deviations from the Gaussian prior, we can preserve prior likelihood exactly by moving along the fixed-radius sphere passing through the initial noise. 
Second, diffusion models are sensitive to the frequency structure of their noise inputs. Empirically, low-frequency noise components provide a more stable subspace for controlling image structure and fidelity, whereas high-frequency updates tend to be less stable and more artifact-prone~\citep{falck2025fourier,harrington2026s,jeon2026lens}. We therefore restrict optimization to the low-frequency subspace while keeping high-frequency components fixed.
This yields a constrained noise manifold, namely an affine sphere over the low-frequency subspace, where prior validity and frequency stability are enforced by construction rather than balanced through auxiliary losses.

Building on this insight, we propose \emph{MoNO}, short for \textbf{M}anifold-c\textbf{o}nstrained \textbf{N}oise \textbf{O}ptimization, a training-free inference-time method that recovers diversity by optimizing initial noises on a constrained manifold. Given the features of previously generated images, MoNO samples a new initial noise, predicts the corresponding clean image with a single network evaluation, and extracts a visual feature vector from the prediction. It then updates the initial noise to reduce the feature vector's projection onto the subspace spanned by previous generations, encouraging the new image to occupy a direction not already covered by the set. The update is performed by Riemannian optimization on the constrained manifold, so the noise remains within the prescribed feasible set throughout optimization~\citep{absil2008optimization,boumal2023introduction}.
This design replaces softly balanced additional terms with hard feasibility constraints. Instead of using auxiliary quality-control objectives to keep the optimized noise from producing degraded images, MoNO constrains every update to remain on the quality-stabilizing noise manifold. Consequently, it can take large geodesic steps without leaving the feasible set, leading to far fewer iterations than prior noise-optimization methods.

Our contributions are as follows:
\begin{itemize}
\item We identify a hard-constrained geometry for noise optimization that preserves Gaussian prior likelihood by fixed-radius norm constraints and stabilizing generation by restricting updates to low-frequency components, yielding a low-frequency affine sphere as the feasible manifold.
\item We propose MoNO, a sequential noise-optimization method that diversifies generations by minimizing each new sample's feature projection onto the subspace spanned by previous samples, using Riemannian updates on the constrained manifold. Because every update remains feasible by construction, MoNO avoids auxiliary quality-control objectives, supports large geodesic steps, and remains training-free and plug-and-play for pretrained few-step generators.
\item We validate MoNO across multiple distilled text-to-image diffusion models. The results show that MoNO consistently improves per-prompt diversity over i.i.d.\ sampling and representative inference-time baselines while keeping comparable image quality.
\end{itemize}

\section{Related Work}

\subsection{Diversity Collapse in Few-Step Diffusion Models}

Few-step distilled diffusion models have made text-to-image generation substantially more efficient, but this acceleration can reduce the effect of sampling randomness. For a fixed prompt, different random seeds may still produce highly similar images, leading to per-prompt diversity collapse~\citep{gandikota2026distilling,wu2026diversity,li20261}. Unlike dataset-level mode dropping, this is a conditional sampling issue. The model may produce realistic images across prompts, but different seeds for the same prompt fail to explore distinct plausible outcomes. Since users often sample several candidates from one prompt to explore alternatives, restoring per-prompt diversity at inference time is especially important for few-step diffusion models.

\subsection{Inference-Time Methods for Diverse Generation}

Existing inference-time methods improve diversity without retraining the model. 
Guidance methods modify the denoising process, either by perturbing the conditioning signal~\citep{sadat2024cads}, introducing interactions among concurrently generated samples~\citep{corso2024particle,vinograd2026diverse}, or applying feature-level adversarial guidance~\citep{singh2024negative}.
Search-based methods leave the sampler unchanged, but generate a large candidate pool and select a subset that balances sample quality and pairwise diversity~\citep{parmar2025scaling}. 
These methods show that diversity can be improved after training, but they either intervene inside the sampling trajectory or rely on generating and filtering extra candidates. For few-step diffusion models, the denoising trajectory offers only a few opportunities for step-wise intervention, while large candidate pools would offset the efficiency gained by distillation. A more direct control variable is therefore the source of stochasticity itself: the initial noise.

\subsection{Noise Optimization for Diverse Generation}

A more direct way to improve generation diversity is to optimize the initial noise itself. Let \(G_\theta\) denote a fixed pretrained diffusion model, and let \(Z=\{z_i\}_{i=1}^N\subset\mathbb{R}^d\) be a batch of initial noises for a prompt \(c\). Existing diversity-oriented noise optimization methods update \(Z\) while keeping \(G_\theta\) fixed, so that the generated samples \(G_\theta(Z,c)\) become more diverse. These methods can be abstracted as
\[
    \min_{Z\in(\mathbb{R}^d)^N}
    \mathcal{L}_{\mathrm{div}}\big(G_\theta(Z,c)\big)
    +
    \lambda\,
    \mathcal{L}_{\mathrm{reg}}(Z,c),
\]
where \(\mathcal{L}_{\mathrm{div}}\) encourages separation among generated samples, and \(\mathcal{L}_{\mathrm{reg}}\) denotes auxiliary regularization used to preserve image fidelity or consistency with the noise prior. 

Different methods instantiate these terms differently. \citet{kim2025diverse} defines a contrastive objective using repulsion between batch elements to increase diversity and attraction to fixed anchors to preserve fidelity. \citet{harrington2026s} optimize initial noises with feature-space diversity objectives and combine them with quality rewards and regularization to avoid quality degradation.
These methods demonstrate that the initial noise is an effective control variable for recovering diversity. However, since \(Z\) is optimized in an unconstrained  Euclidean space, fidelity and prior preservation must be enforced indirectly through soft regularization terms or optimization heuristics. As a result, diversity improvement must be balanced against fidelity through objective weights, stopping rules, or conservative update sizes. This suggests that the admissible region of noise optimization is as important as the objective used to encourage diversity.

\section{Method}

\begin{figure}[t]
  \centering
  \includegraphics[width=\linewidth]{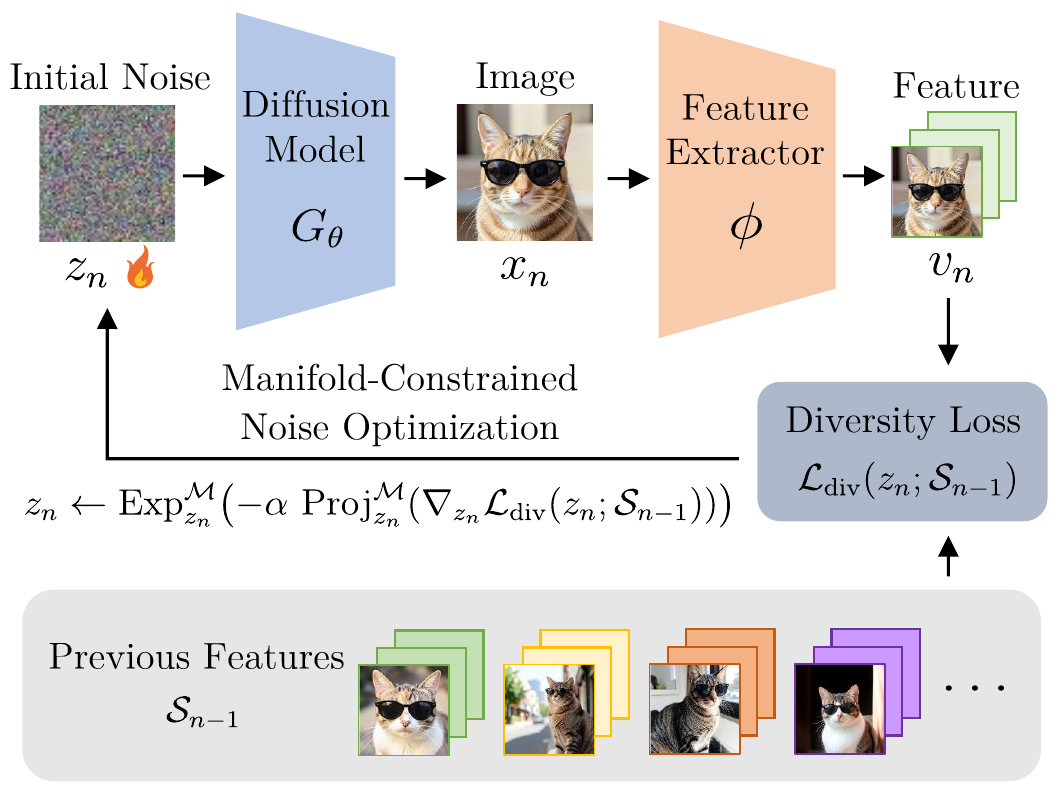}
    \caption{
    Overview of MoNO. Given a new initial noise \(z_n\), the diffusion model produces an image \(x_n\), whose feature \(v_n\) is compared with the previous features \(\mathcal{S}_{n-1}\). The resulting diversity loss drives manifold-constrained noise optimization, updating only the initial noise while keeping the model parameters and sampler fixed.
    }
  \label{fig:method_overview}
\end{figure}

This section presents the proposed MoNO framework. We first formulate diversity enhancement as a sequential feature-subspace optimization problem. We then define the admissible noise manifold that constrains the optimization, derive the corresponding Riemannian update, and finally describe the practical instantiation used for efficient inference.
\Cref{fig:method_overview} provides an overview of the full pipeline.

\subsection{Sequential Feature-Subspace Diversification}

Let \(c\) denote a text prompt, and let \(x = G_\theta(z,c)\) be the image generated by a fixed pretrained diffusion model from an initial noise \(z\in\mathbb{R}^d\). Given a frozen feature extractor \(\phi\), we write the feature of the generated image as
\[
    v(z,c)=\phi(x)=\phi(G_\theta(z,c)).
\]
Following the sequential sampling strategy of~\citet{harrington2026s}, MoNO constructs a diverse set one sample at a time, which avoids optimizing all samples jointly and allows the set size to grow without a fixed batch-size limit. Suppose that \(n-1\) samples have already been generated for the same prompt, with initial noises \(\{z_i\}_{i=1}^{n-1}\) and corresponding features
\[
    v_i = v(z_i,c), \quad i=1,\ldots,n-1 .
\]
These previous features define the occupied feature subspace
\[
    \mathcal{S}_{n-1}
    =
    \operatorname{span}(v_1,\ldots,v_{n-1}).
\]

For the \(n\)-th sample, MoNO starts from a fresh Gaussian noise \(z_n^0\sim\mathcal{N}(0,I)\) and optimizes the noise so that the new feature is not well explained by \(\mathcal{S}_{n-1}\). Let \(\Pi_{\mathcal{S}_{n-1}}\) denote the orthogonal projector onto \(\mathcal{S}_{n-1}\). We define the sequential diversity loss as
\[
    \mathcal{L}_{\mathrm{div}}(z_n;\mathcal{S}_{n-1})
    =
    \left\|
    \Pi_{\mathcal{S}_{n-1}}
    v(z_n,c)
    \right\|_2^2 .
\]
Minimizing this loss suppresses the component of the new feature that lies in the subspace spanned by previous generations. 
The \(n\)-th noise is therefore obtained by solving
\[
    z_n^\star
    =
    \arg\min_{z_n\in\mathcal{M}_n}
    \mathcal{L}_{\mathrm{div}}(z_n;\mathcal{S}_{n-1}),
\]
where \(\mathcal{M}_n\subset\mathbb{R}^d\) denotes the admissible noise set. This formulation differs from noise optimization with an auxiliary soft regularizer because fidelity-preserving constraints are imposed through the feasible set itself, rather than added as a separate regularization term. After obtaining \(z_n^\star\), the generated sample \(G_\theta(z_n^\star,c)\) is added to the set, and its feature is incorporated into the feature subspace for subsequent sequential updates.

\subsection{Manifold-Constrained Noise Space}

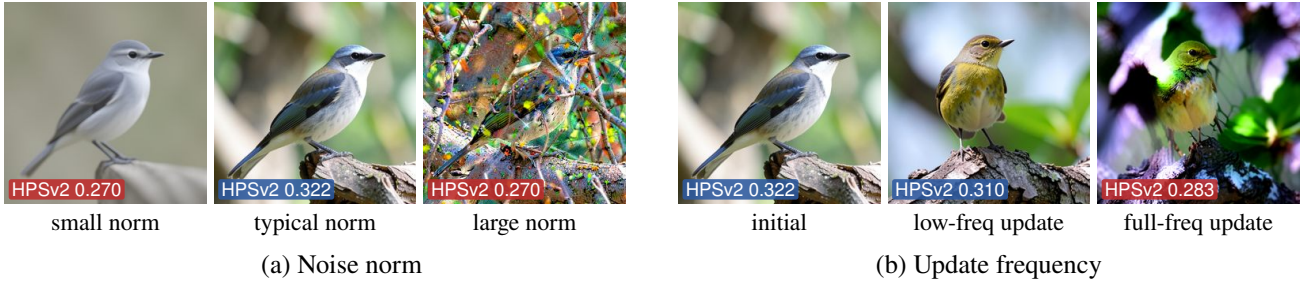
\begin{figure*}[t]
  \centering
  \setlength{\tabcolsep}{1.5pt}
  \newcommand{\imgw}{0.15\linewidth}
  \definecolor{hpsbad}{RGB}{190,45,45}
  \definecolor{hpsok}{RGB}{45,90,160}

  \newcommand{\imgbadge}[3][hpsok]{%
    \begin{tikzpicture}
      \node[inner sep=0] (im) {\includegraphics[width=\imgw]{#2}};
      \node[anchor=south west, inner sep=1.2pt, font=\scriptsize\sffamily,
            text=white, fill=#1, fill opacity=0.88, text opacity=1,
            rounded corners=1pt]
           at ([shift={(1.5pt,1.5pt)}]im.south west) {HPSv2~#3};
    \end{tikzpicture}}

  \begin{tabular}{@{}ccc@{\hspace{20pt}}ccc@{}}
    \imgbadge[hpsbad]{method/norm/low.jpg}{0.270}      &
    \imgbadge{method/norm/std.jpg}{0.322}              &
    \imgbadge[hpsbad]{method/norm/high.jpg}{0.270}     &
    \imgbadge{method/frequency/ref.jpg}{0.322}         &
    \imgbadge{method/frequency/low.jpg}{0.310}         &
    \imgbadge[hpsbad]{method/frequency/all.jpg}{0.283} \\[-1pt]
    {\small small norm} & {\small typical norm} & {\small large norm} &
    {\small initial} & {\small low-freq update} & {\small full-freq update} \\[5pt]
    \multicolumn{3}{c}{(a) Noise norm} &
    \multicolumn{3}{c}{(b) Update frequency}
  \end{tabular}

  \caption{
  Effects of noise norm and update frequency on generation quality for the prompt \textit{``a photo of a bird''}. Badge colors indicate relative visual quality.
  (a) Initial noises with different norms produce images with visibly different fidelity.
  (b) Starting from the same initial generation, a one-step norm-preserving update restricted to low frequencies is less disruptive than a one-step norm-preserving update over all frequency components.
  }
  \label{fig:noise_constraints}
\end{figure*}

The sequential diversity loss specifies which feature component should be suppressed. We now define where the initial noise is allowed to move during optimization. Instead of adding a soft regularization term to penalize undesirable noise updates, MoNO imposes fidelity-preserving constraints through the feasible set itself. \Cref{fig:noise_constraints} illustrates the two empirical observations behind this design: generated images are sensitive to the norm of the initial noise, and full-frequency updates can be substantially more disruptive than low-frequency updates.

\paragraph{Norm constraint.}
The initial noise follows the standard Gaussian prior \(z\sim\mathcal{N}(0,I)\), whose density depends only on the Euclidean norm \(\|z\|_2\). An unconstrained update can therefore change the prior likelihood by moving the noise radially. To avoid this radial drift, MoNO constrains the optimized noise to remain on the sphere determined by its initial draw \(z^0\):
\[
    \mathcal{M}_{\mathrm{norm}}(z^0)
    =
    \left\{
    z\in\mathbb{R}^d:
    \|z\|_2=\|z^0\|_2
    \right\}.
\]
This constraint allows the noise to move along a fixed-radius Gaussian shell rather than toward a different-norm region. As shown in \Cref{fig:noise_constraints}(a), noises with different norms can lead to markedly different generation quality.

\paragraph{Frequency constraint.}
A fixed norm alone is not sufficient. Although spherical updates preserve the Gaussian density, they may still modify high-frequency components of the initial noise. Prior studies suggest that diffusion models expose meaningful and uneven control across noise frequencies, with low-frequency components often providing a more stable handle for image structure and fidelity~\citep{falck2025fourier,harrington2026s,jeon2026lens}. We observe the same tendency in \Cref{fig:noise_constraints}(b): under a one-step norm-preserving update, allowing all frequency components to change is more disruptive than restricting the update to low frequencies. We therefore constrain the optimization trajectory to a low-frequency subspace. Let \(P_L\) denote an orthogonal low-pass projection and let \(P_H=I-P_L\) be its high-frequency complement. MoNO freezes the high-frequency component of the initial noise:
\[
    \mathcal{M}_{\mathrm{freq}}(z^0)
    =
    \left\{
    z\in\mathbb{R}^d:
    P_H z = P_H z^0
    \right\}.
\]
This constraint ensures that only the low-frequency component of the noise can be changed during optimization.

\paragraph{Intersection manifold.}
Combining the norm and frequency constraints yields the admissible noise manifold:
\[
    \mathcal{M}(z^0)
    =
    \mathcal{M}_{\mathrm{norm}}(z^0)
    \cap
    \mathcal{M}_{\mathrm{freq}}(z^0).
\]
Equivalently, decompose the initial noise as
\[
    z^0 = z_L^0 + z_H^0,
    \quad
    z_L^0=P_Lz^0,\quad z_H^0=P_Hz^0 .
\]
Since \(P_L\) and \(P_H\) are orthogonal projections, the feasible set can be written as an affine low-frequency sphere:
\[
    \mathcal{M}(z^0)
    =
    \left\{
    z=z_H^0+u:
    u\in\operatorname{Im}(P_L),
    \;
    \|u\|_2=\|z_L^0\|_2
    \right\}.
\]
Thus, MoNO optimizes only the low-frequency component, keeps the low-frequency norm fixed, and leaves the high-frequency component unchanged. This converts fidelity preservation from a soft regularization term into a hard geometric constraint on the initial-noise space.

\subsection{Riemannian Noise Update}

Because \(\mathcal{M}(z^0)\) is a low-dimensional manifold, ordinary gradient updates do not preserve feasibility. We therefore use the Riemannian gradient descent recipe~\citep{boumal2023introduction}. Fixing the current sequential step \(n\), we write the objective as
\[
    f(z)=\mathcal{L}_{\mathrm{div}}(z).
\]
From the previous subsection, any feasible noise can be represented as
\[
    z = z_H^0 + u,\quad
    \|u\|_2 = r,
\]
where \(z_H^0=P_Hz^0\), \(u=P_Lz\), and \(r=\|P_Lz^0\|_2\). Thus, the feasible set is a sphere in the low-frequency subspace, translated by the fixed high-frequency component \(z_H^0\).

At iteration \(k\), let
\[
    z^k=z_H^0+u^k,\quad \|u^k\|_2=r .
\]
We first compute the Euclidean gradient in the ambient noise space,
\[
    g^k=\nabla_z f(z^k).
\]
Feasible infinitesimal updates must keep the high-frequency component fixed and must not change the norm of the low-frequency component. Hence, the descent direction should lie in the tangent space of the feasible manifold. The tangent space of \(\mathcal{M}(z^0)\) at \(z^k\) is
\[
    T_{z^k}\mathcal{M}(z^0)
    =
    \left\{
    \xi\in\operatorname{Im}(P_L):
    \langle \xi,u^k\rangle = 0
    \right\}.
\]
The orthogonal projection of \(g^k\) onto this tangent space is
\[
    \operatorname{Proj}_{z^k}^{\mathcal{M}(z^0)}(g^k)
    =
    P_Lg^k
    -
    \frac{\langle P_Lg^k,u^k\rangle}{\|u^k\|_2^2}u^k .
\]
This gives the Riemannian gradient
\[
    \operatorname{grad} f(z^k)
    =
    \operatorname{Proj}_{z^k}^{\mathcal{M}(z^0)}(g^k).
\]
The projection removes both infeasible components of the Euclidean gradient: the high-frequency component, which would change \(P_Hz\), and the radial component, which would change the low-frequency norm.

We then take the Riemannian descent direction
\[
    h^k=-\operatorname{grad} f(z^k).
\]
With step size \(\alpha>0\), the next iterate is obtained by moving along the geodesic on the manifold, equivalently by the exponential map:
\[
    z^{k+1}
    =
    \operatorname{Exp}_{z^k}^{\mathcal{M}(z^0)}
    \left(
    \alpha h^k
    \right).
\]
Since \(\mathcal{M}(z^0)\) is an affine sphere, this geodesic update has a closed form. With \(z^k=z_H^0+u^k\), the low-frequency component is updated by
\[
    u^{k+1}
    =
    u^k
    \cos\left(\frac{\alpha\|h^k\|_2}{r}\right)
    +
    r\frac{h^k}{\|h^k\|_2}
    \sin\left(\frac{\alpha\|h^k\|_2}{r}\right),
\]
and the full noise is reconstructed as
\[
    z^{k+1}=z_H^0+u^{k+1}.
\]
By construction, the update satisfies
\[
    P_Hz^{k+1}=z_H^0,\quad
    \|P_Lz^{k+1}\|_2=r,
\]
so each iterate remains exactly on \(\mathcal{M}(z^0)\).

\begin{table*}[t]
\centering
\setlength{\tabcolsep}{4pt}    
\renewcommand{\arraystretch}{1.12}
\caption{
Main results on GenEval with distilled text-to-image diffusion models.
We report the mean across prompts, with the standard deviation shown in subscript.
The best results within each model block are shown in bold.
}
\label{tab:main_geneval}
\begin{tabular}{c|l |ccc | ccc}
\toprule
\multirow{2}{*}{Model}
& \multirow{2}{*}{Method}
& \multicolumn{3}{c|}{Diversity $\uparrow$}
& \multicolumn{3}{c}{Quality $\uparrow$} \\
\cmidrule(lr){3-5}
\cmidrule(lr){6-8}
& & LPIPS & DreamSim & Vendi Score
& CLIPScore & CLIP-IQA & HPSv2 \\
\midrule
\multirow{9}{*}{\shortstack[c]{FLUX.2 \\[3pt] {[klein]} 4B \\[3pt] (4 steps)}}
& i.i.d.
& \ms{0.528}{0.133} & \ms{0.178}{0.078} & \ms{1.734}{0.461}
& \bms{0.338}{0.035} & \ms{0.630}{0.112} & \ms{0.313}{0.025} \\
& CADS
& \ms{0.733}{0.040} & \ms{0.413}{0.097} & \ms{2.575}{0.608}
& \ms{0.323}{0.031} & \ms{0.558}{0.118} & \ms{0.288}{0.025} \\
& Particle Guidance
& \ms{0.540}{0.123} & \ms{0.183}{0.078} & \ms{1.757}{0.468}
& \bms{0.338}{0.034} & \bms{0.636}{0.118} & \ms{0.313}{0.025} \\
& NegToMe
& \ms{0.544}{0.123} & \ms{0.185}{0.079} & \ms{1.771}{0.465}
& \bms{0.338}{0.034} & \ms{0.631}{0.118} & \ms{0.312}{0.025} \\
& Group Inference
& \ms{0.609}{0.101} & \ms{0.232}{0.092} & \ms{1.934}{0.536}
& \ms{0.337}{0.033} & \ms{0.631}{0.119} & \ms{0.311}{0.024} \\
& DivGen
& \ms{0.553}{0.114} & \ms{0.191}{0.081} & \ms{1.809}{0.495}
& \bms{0.338}{0.034} & \bms{0.636}{0.121} & \bms{0.315}{0.026} \\
& CNO
& \ms{0.540}{0.123} & \ms{0.183}{0.078} & \ms{1.761}{0.468}
& \bms{0.338}{0.034} & \bms{0.636}{0.118} & \ms{0.313}{0.025} \\
\rowcolor{oursgray}
\cellcolor{white} & MoNO (1-iter)
& \ms{0.692}{0.072} & \ms{0.330}{0.103} & \ms{2.168}{0.547}
& \ms{0.336}{0.032} & \ms{0.613}{0.111} & \ms{0.303}{0.024} \\
\rowcolor{oursgray}
\cellcolor{white} & MoNO (4-iter)
& \bms{0.742}{0.064} & \bms{0.453}{0.103} & \bms{2.592}{0.599}
& \ms{0.332}{0.032} & \ms{0.610}{0.109} & \ms{0.294}{0.023} \\
\midrule
& i.i.d.
& \ms{0.628}{0.054} & \ms{0.234}{0.091} & \ms{1.937}{0.508}
& \bms{0.330}{0.030} & \ms{0.605}{0.121} & \ms{0.297}{0.031} \\
& DivGen
& \ms{0.650}{0.057} & \ms{0.268}{0.093} & \ms{2.161}{0.557}
& \ms{0.329}{0.030} & \ms{0.576}{0.122} & \ms{0.294}{0.031} \\
& CNO
& \ms{0.643}{0.058} & \ms{0.247}{0.090} & \ms{1.975}{0.497}
& \bms{0.330}{0.030} & \ms{0.574}{0.123} & \ms{0.295}{0.031} \\
\rowcolor{oursgray}
\cellcolor{white} & MoNO (1-iter)
& \ms{0.695}{0.049} & \ms{0.320}{0.104} & \ms{2.162}{0.547}
& \ms{0.328}{0.030} & \ms{0.608}{0.116} & \bms{0.299}{0.028} \\
\rowcolor{oursgray}
\cellcolor{white}\multirow{-5}{*}{\shortstack[c]{SDXL-Turbo \\[3pt] (1 step)}} & MoNO (4-iter)
& \bms{0.735}{0.049} & \bms{0.384}{0.107} & \bms{2.368}{0.563}
& \ms{0.326}{0.030} & \bms{0.619}{0.111} & \ms{0.297}{0.027} \\
\midrule
& i.i.d.
& \ms{0.646}{0.084} & \ms{0.257}{0.081} & \ms{2.037}{0.483}
& \bms{0.333}{0.029} & \ms{0.606}{0.128} & \ms{0.294}{0.030} \\
& DivGen
& \ms{0.680}{0.076} & \ms{0.288}{0.088} & \ms{2.312}{0.549}
& \ms{0.331}{0.030} & \ms{0.583}{0.131} & \bms{0.295}{0.029} \\
& CNO
& \ms{0.674}{0.076} & \ms{0.267}{0.083} & \ms{2.068}{0.497}
& \ms{0.332}{0.030} & \ms{0.580}{0.134} & \bms{0.295}{0.030} \\
\rowcolor{oursgray}
\cellcolor{white} & MoNO (1-iter)
& \ms{0.733}{0.062} & \ms{0.338}{0.087} & \ms{2.283}{0.546}
& \ms{0.330}{0.029} & \ms{0.606}{0.119} & \ms{0.294}{0.028} \\
\rowcolor{oursgray}
\cellcolor{white}\multirow{-5}{*}{\shortstack[c]{DMD2 \\[3pt] (1 step)}} & MoNO (4-iter)
& \bms{0.768}{0.053} & \bms{0.385}{0.090} & \bms{2.493}{0.554}
& \ms{0.327}{0.028} & \bms{0.610}{0.107} & \ms{0.293}{0.025} \\
\bottomrule
\end{tabular}
\end{table*}

\begin{figure*}[t]
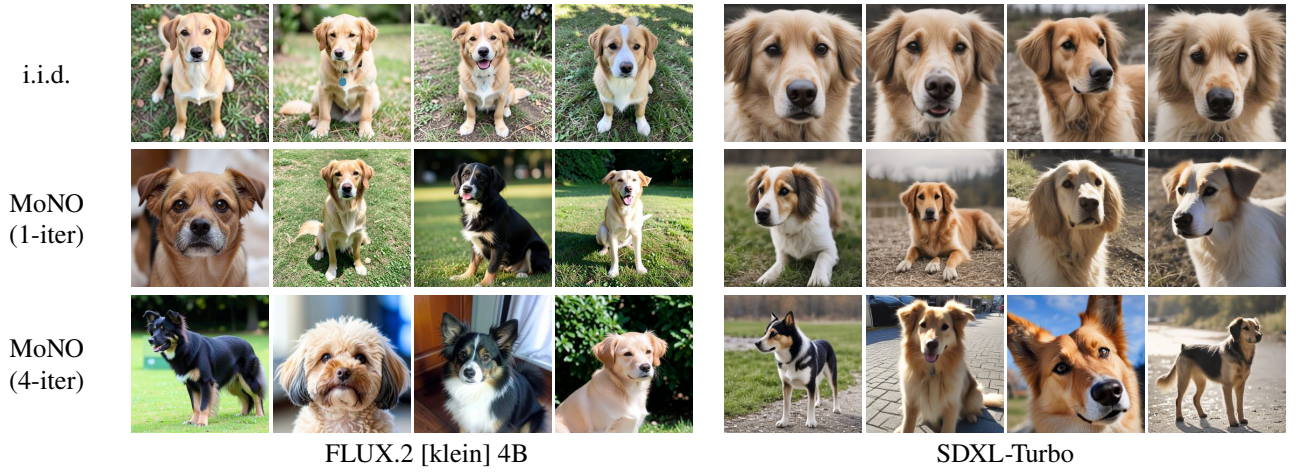

\centering
\setlength{\tabcolsep}{0.8pt}
\renewcommand{\arraystretch}{1.0}
\newcommand{\mainimg}[1]{\includegraphics[width=\linewidth]{#1}}
\begin{tabular}{@{}
  >{\centering\arraybackslash}m{0.12\textwidth}
  *{4}{>{\centering\arraybackslash}m{0.102\textwidth}}
  @{\hspace{12pt}}
  *{4}{>{\centering\arraybackslash}m{0.102\textwidth}}@{}}
i.i.d.
& \mainimg{experiments/main/flux_klein/iid/1.jpg}
& \mainimg{experiments/main/flux_klein/iid/2.jpg}
& \mainimg{experiments/main/flux_klein/iid/3.jpg}
& \mainimg{experiments/main/flux_klein/iid/4.jpg}
& \mainimg{experiments/main/sdxl_turbo/iid/1.jpg}
& \mainimg{experiments/main/sdxl_turbo/iid/2.jpg}
& \mainimg{experiments/main/sdxl_turbo/iid/3.jpg}
& \mainimg{experiments/main/sdxl_turbo/iid/4.jpg} \\[2pt]
\shortstack{MoNO\\[2pt](1-iter)}
& \mainimg{experiments/main/flux_klein/iter1/1.jpg}
& \mainimg{experiments/main/flux_klein/iter1/2.jpg}
& \mainimg{experiments/main/flux_klein/iter1/3.jpg}
& \mainimg{experiments/main/flux_klein/iter1/4.jpg}
& \mainimg{experiments/main/sdxl_turbo/iter1/1.jpg}
& \mainimg{experiments/main/sdxl_turbo/iter1/2.jpg}
& \mainimg{experiments/main/sdxl_turbo/iter1/3.jpg}
& \mainimg{experiments/main/sdxl_turbo/iter1/4.jpg} \\[2pt]
\shortstack{MoNO\\[2pt](4-iter)}
& \mainimg{experiments/main/flux_klein/iter4/1.jpg}
& \mainimg{experiments/main/flux_klein/iter4/2.jpg}
& \mainimg{experiments/main/flux_klein/iter4/3.jpg}
& \mainimg{experiments/main/flux_klein/iter4/4.jpg}
& \mainimg{experiments/main/sdxl_turbo/iter4/1.jpg}
& \mainimg{experiments/main/sdxl_turbo/iter4/2.jpg}
& \mainimg{experiments/main/sdxl_turbo/iter4/3.jpg}
& \mainimg{experiments/main/sdxl_turbo/iter4/4.jpg} \\[4pt]
& \multicolumn{4}{c}{FLUX.2 [klein] 4B}
& \multicolumn{4}{c}{SDXL-Turbo} \\
\end{tabular}
\caption{
Visual comparison of per-prompt generation diversity on FLUX.2 [klein] 4B and SDXL-Turbo for the prompt \textit{``a photo of a dog''}.
Standard i.i.d.\ sampling produces visually similar samples, while MoNO increases diversity with one optimization iteration and yields substantially more diverse generations with four iterations.
}
\label{fig:main_qualitative}
\end{figure*}

Compared with a free Euclidean update, the Riemannian update first discards the gradient components that would violate the constraints. It then moves along the geodesic of \(\mathcal{M}(z^0)\), producing a constraint-preserving update of the initial noise. Thus, every iterate remains feasible without auxiliary regularization or conservative post-hoc correction.
Additional derivation details are provided in the appendix.

\subsection{Practical Instantiation}

In the formulation above, the diversity loss is written with the generator output \(x=G_\theta(z,c)\) and the feature subspace is defined abstractly by previous features. In practice, we instantiate both components in a lightweight way. 
To avoid differentiating through the full sampling trajectory, we compute the optimization loss using the denoising network's own single-step clean-image prediction, denoted by \(A_\theta\). Given the current noise \(z\), this proxy prediction produces
\(
    \hat{x}=A_\theta(z,c).
\)
During optimization, we use \(v(z,c)=\phi(\hat{x})\) for gradient computation. After optimization, the final image is still generated as \(x=G_\theta(z,c)\) using the original sampler.
We instantiate \(\phi\) with lightweight DINOv2 patch features~\citep{oquab2023dinov2}.
We also maintain an orthonormal basis \(Q_{n-1}\) of the previous feature subspace \(\mathcal{S}_{n-1}\), so that the projection loss is computed as
\[
    \mathcal{L}_{\mathrm{div}}(z_n;\mathcal{S}_{n-1})
    =
    \left\|
    Q_{n-1}^{\top}v(z_n,c)
    \right\|_2^2 .
\]

\Cref{alg:mono} summarizes the full procedure. MoNO generates samples sequentially, optimizing each newly sampled noise on its own manifold \(\mathcal{M}(z_n^0)\) and then adding the resulting feature to the maintained basis. The first sample is generated from an empty basis, while later samples are optimized to avoid the subspace already covered by previous generations. Since MoNO stores only a compact feature basis and optimizes one noise at a time, its memory cost is effectively independent of the final set size, and the number of generated samples can be chosen freely at inference time.

\begin{algorithm}[t]
\caption{Manifold-Constrained Noise Optimization}
\label{alg:mono}
\renewcommand{\arraystretch}{1.18}
\begin{algorithmic}[1]
\Require prompt \(c\), generator \(G_\theta\), proxy predictor \(A_\theta\), feature extractor \(\phi\), low-pass projection \(P_L\), number of samples \(N\), optimization iterations \(K\), step size \(\alpha\)
\State Initialize \(Q_0\leftarrow [\,]\)
\For{\(n=1,\ldots,N\)}
    \State Sample \(z_n^0\sim\mathcal{N}(0,I)\)
    \State Construct \(\mathcal{M}(z_n^0)\) using \(P_L\) and \(P_H=I-P_L\)
    \State Set \(z_n^0\) as the initial iterate
    \For{\(k=0,\ldots,K-1\)}
        \State \(\hat{x}\leftarrow A_\theta(z_n^k,c)\)
        \State \(v\leftarrow \phi(\hat{x})\)
        \State \(f(z_n^k)\leftarrow \|Q_{n-1}^{\top}v\|_2^2\)
        \State \(g^k\leftarrow \nabla_{z_n^k}f(z_n^k)\)
        \State \(\operatorname{grad} f(z_n^k)\leftarrow \operatorname{Proj}_{z_n^k}^{\mathcal{M}(z_n^0)}(g^k)\)
        \State \(z_n^{k+1}\leftarrow \operatorname{Exp}_{z_n^k}^{\mathcal{M}(z_n^0)}\!\left(-\alpha\,\operatorname{grad} f(z_n^k)\right)\)
    \EndFor
    \State Generate \(x_n\leftarrow G_\theta(z_n^K,c)\)
    \State Extract \(v_n\leftarrow \phi(x_n)\)
    \State \(Q_n\leftarrow\operatorname{Orthonormalize}([Q_{n-1},v_n])\)
\EndFor
\State \Return \(\{x_n\}_{n=1}^N\)
\end{algorithmic}
\end{algorithm}

\section{Experiments}

\begin{figure}[t]
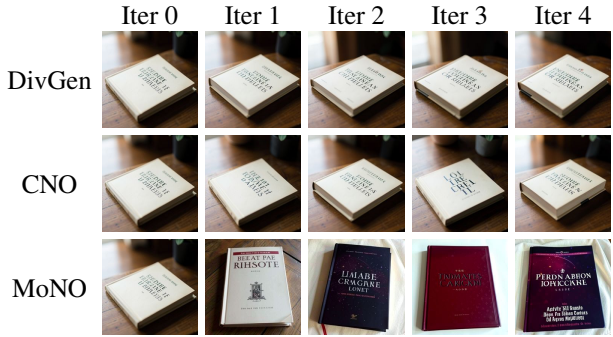

  \centering
  \setlength{\tabcolsep}{1.5pt}
  \renewcommand{\arraystretch}{0.95}

  \newcommand{\iterimg}[1]{%
    \includegraphics[width=0.15\columnwidth]{#1}%
  }

  \begin{tabular}{@{}
    >{\centering\arraybackslash}m{0.14\columnwidth}
    *{5}{>{\centering\arraybackslash}m{0.15\columnwidth}}@{}}
    &  Iter 0
    &  Iter 1
    &  Iter 2
    &  Iter 3
    &  Iter 4 \\

    DivGen
    & \iterimg{experiments/comparison/divgen/0.jpg}
    & \iterimg{experiments/comparison/divgen/1.jpg}
    & \iterimg{experiments/comparison/divgen/2.jpg}
    & \iterimg{experiments/comparison/divgen/3.jpg}
    & \iterimg{experiments/comparison/divgen/4.jpg} \\

    CNO
    & \iterimg{experiments/comparison/cno/0.jpg}
    & \iterimg{experiments/comparison/cno/1.jpg}
    & \iterimg{experiments/comparison/cno/2.jpg}
    & \iterimg{experiments/comparison/cno/3.jpg}
    & \iterimg{experiments/comparison/cno/4.jpg} \\

    MoNO
    & \iterimg{experiments/comparison/mono/0.jpg}
    & \iterimg{experiments/comparison/mono/1.jpg}
    & \iterimg{experiments/comparison/mono/2.jpg}
    & \iterimg{experiments/comparison/mono/3.jpg}
    & \iterimg{experiments/comparison/mono/4.jpg}
  \end{tabular}

  \caption{
  Optimization progress of three different noise-optimization methods on FLUX.2 [klein] 4B for the prompt \textit{``a photo of a book''}. MoNO yields more visible diversification after only a few updates.
  }
  \label{fig:optimization_iterations}
\end{figure}

\begin{table}[t]
\centering
\setlength{\tabcolsep}{2.5pt}
\renewcommand{\arraystretch}{1.12}
\caption{
Ablation of MoNO constraints on GenEval with FLUX.2 [klein] 4B.
All variants use four optimization iterations with the same update magnitude.
The best results are shown in bold.
}
\label{tab:ablation_constraints}
\begin{tabular}{lccc}
\toprule
Constraints & CLIPScore $\uparrow$ & CLIP-IQA $\uparrow$ & HPSv2 $\uparrow$ \\
\midrule
Norm + Freq
& \bms{0.332}{0.032}
& \bms{0.610}{0.109}
& \bms{0.294}{0.023} \\
Norm only
& \ms{0.301}{0.025}
& \ms{0.490}{0.096}
& \ms{0.255}{0.020} \\
Freq only
& \ms{0.329}{0.030}
& \ms{0.605}{0.109}
& \ms{0.290}{0.023} \\
\bottomrule
\end{tabular}
\end{table}

\begin{table}[t]
\centering
\setlength{\tabcolsep}{5pt}
\renewcommand{\arraystretch}{1.12}
\caption{
Ablation of the single-step proxy predictor on GenEval with FLUX.2 [klein] 4B.
Both variants use four optimization iterations.
The best results are shown in bold.
}
\label{tab:ablation_proxy}
\begin{tabular}{c|l|cc}
\toprule
\multicolumn{2}{c|}{Metric} & Full  & Proxy  \\
\midrule
\multirow{3}{*}{Diversity $\uparrow$}
& LPIPS
& \ms{0.724}{0.064}
& \bms{0.742}{0.064} \\
& DreamSim
& \ms{0.423}{0.104}
& \bms{0.453}{0.103} \\
& Vendi Score
& \bms{2.593}{0.572}
& \ms{2.592}{0.599} \\
\midrule
\multirow{3}{*}{Quality $\uparrow$}
& CLIPScore
& \ms{0.331}{0.032}
& \bms{0.332}{0.032} \\
& CLIP-IQA
& \bms{0.632}{0.104}
& \ms{0.610}{0.109} \\
& HPSv2
& \bms{0.301}{0.024}
& \ms{0.294}{0.023} \\
\bottomrule
\end{tabular}
\end{table}

\subsection{Experimental Setup}

\paragraph{Models and datasets.}
In the main experiments, we evaluate MoNO on few-step distilled text-to-image diffusion models, including 4-step FLUX.2 [klein] 4B~\citep{flux-2-2025}, single-step SDXL-Turbo~\citep{sauer2024adversarial}, and single-step DMD2~\citep{yin2024improved}.
We evaluate all methods on the GenEval benchmark~\citep{ghosh2023geneval}, generating four images for each prompt to assess per-prompt generation diversity.
Unless otherwise specified, all other sampling parameters are kept at their default values.

\paragraph{Compared methods.}
In addition to standard i.i.d. sampling, we compare MoNO with representative inference-time diversity sampling methods, including CADS~\citep{sadat2024cads}, Particle Guidance~\citep{corso2024particle}, NegToMe~\citep{singh2024negative}, Group Inference~\citep{parmar2025scaling}, DivGen~\citep{harrington2026s}, and CNO~\citep{kim2025diverse}. 
For the single-step distilled model, step-wise methods are less directly applicable, and we therefore compare only with noise optimization methods. 
For all noise optimization methods, we use four optimization iterations in the main comparison. 
For MoNO, we additionally report the one-iteration variant.

\paragraph{Evaluation metrics.}
We evaluate each method in terms of diversity and quality. 
We measure diversity using pairwise LPIPS~\citep{zhang2018unreasonable}, pairwise DreamSim~\citep{fu2023dreamsim}, and the set-level Vendi Score~\citep{friedman2022vendi}. 
We measure quality using per-image CLIPScore~\citep{hessel2021clipscore}, CLIP-IQA~\citep{wang2023exploring}, and HPSv2~\citep{wu2023human}. 
More detailed experimental settings are provided in the appendix.

\subsection{Main Results}

As shown in \Cref{tab:main_geneval}, MoNO consistently improves generation diversity across all evaluated distilled text-to-image generators while keeping image quality stable. 
Compared with i.i.d.\ sampling, MoNO substantially increases all diversity metrics, indicating that optimizing each new initial noise against the feature subspace of previous generations effectively reduces repeated modes within the same prompt. 
Compared with prior noise-optimization methods, MoNO also achieves stronger diversity gains, and the one-iteration variant already provides a clear improvement, suggesting that the constrained update can make effective progress with very limited optimization.

The visual comparisons in \Cref{fig:main_qualitative} further illustrate the diversity improvements. 
While i.i.d.\ sampling often produces visually similar samples for the same prompt, MoNO generates more varied visual attributes, with four iterations producing stronger diversity than one iteration. 
To inspect the optimization dynamics more directly, \Cref{fig:optimization_iterations} compares different noise-optimization methods across iterations. 
Prior methods show only limited changes, whereas MoNO rapidly diversifies the generated samples within a few iterations, illustrating the benefit of large geodesic updates on the constrained noise manifold in the low-iteration regime.
Additional experimental results are provided in the Appendix.

\subsection{Ablation Study}

We ablate two key design choices in MoNO: the optimization constraints that preserve image quality, and the single-step proxy predictor that enables efficient gradient computation.

\begin{figure}[t]
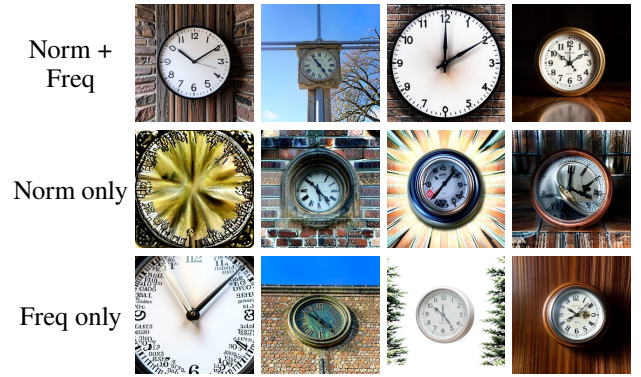

\centering
\setlength{\tabcolsep}{1.5pt}
\renewcommand{\arraystretch}{0.95}
\newcommand{\ablimg}[1]{\includegraphics[width=0.185\columnwidth]{#1}}
\begin{tabular}{@{}
  >{\centering\arraybackslash}m{0.18\columnwidth}
  *{4}{>{\centering\arraybackslash}m{0.185\columnwidth}}@{}}
Norm + Freq
& \ablimg{experiments/ablation/both/1.jpg}
& \ablimg{experiments/ablation/both/2.jpg}
& \ablimg{experiments/ablation/both/3.jpg}
& \ablimg{experiments/ablation/both/4.jpg} \\[2pt]

Norm only
& \ablimg{experiments/ablation/norm/1.jpg}
& \ablimg{experiments/ablation/norm/2.jpg}
& \ablimg{experiments/ablation/norm/3.jpg}
& \ablimg{experiments/ablation/norm/4.jpg} \\[2pt]

Freq only
& \ablimg{experiments/ablation/freq/1.jpg}
& \ablimg{experiments/ablation/freq/2.jpg}
& \ablimg{experiments/ablation/freq/3.jpg}
& \ablimg{experiments/ablation/freq/4.jpg} \\
\end{tabular}
\caption{
Visual comparison of MoNO constraint ablations on the prompt \textit{``a photo of a clock''}.
Using both constraints results high-quality images, while keeping only one constraint causes significant quality degradation.
}
\label{fig:ablation_constraints}
\end{figure}

\paragraph{Effect of optimization constraints.}
We ablate the two constraints used in MoNO: norm constraint and frequency constraint. 
As shown in~\Cref{tab:ablation_constraints}, using either constraint alone leads to a degradation in image quality. 
As illustrated in~\Cref{fig:ablation_constraints}, the norm-only variant suffers from severe high-frequency artifacts and distorted structures, while the frequency-only variant often produces locally over-saturated regions. 
In contrast, combining the norm constraint and the frequency constraint better preserves visual quality while enabling effective diversity enhancement.

\paragraph{Effect of the single-step proxy predictor.}
MoNO optimizes the diversity loss using the denoising network's single-step clean-image prediction, i.e.,
\(\phi(A_\theta(z,c))\), rather than differentiating through the full sampling trajectory, i.e.,
\(\phi(G_\theta(z,c))\).
To validate this design, we compare the default proxy-based optimization with a variant that replaces \(\phi(A_\theta(z,c))\) by \(\phi(G_\theta(z,c))\) in the diversity loss and backpropagates through the full sampler.
As shown in Table~\ref{tab:ablation_proxy}, the proxy-based variant achieves a comparable quality--diversity trade-off to the full-sampler variant, with a measured \(2.37\times\) speedup.
This shows that \(A_\theta\) provides an effective and efficient optimization signal for diverse generation.

\section{Conclusion}

We introduced MoNO, a manifold-constrained noise optimization method for improving per-prompt diversity in few-step distilled text-to-image diffusion models. 
MoNO optimizes each initial noise against the feature subspace of previous generations, while restricting updates to a low-dimensional manifold that preserves the Gaussian prior geometry and fixes unstable high-frequency components. 
This constrained formulation enables larger stable updates and yields consistent diversity gains over i.i.d.\ sampling and prior inference-time baselines, while maintaining image quality.

\appendix


\bibliography{aaai2027}

\clearpage

\setcounter{secnumdepth}{2}
\twocolumn[{%
    \vbox{%
      \hsize\textwidth%
      \linewidth\hsize%
      \vskip 0.625in minus 0.125in%
      \centering%
      {\LARGE\bf Appendix \par}%
      \vskip 3em plus 2fil%
    }%
  }]%
\section{Details of the Riemannian Noise Update}
\label{app:riemannian_update}

This section provides the derivation of the Riemannian update used in MoNO. We use the Euclidean metric inherited from the ambient noise space. Let \(P_L\) be an orthogonal low-pass projection and \(P_H=I-P_L\) be its orthogonal complement. Thus, \(P_L^2=P_L\), \(P_H^2=P_H\), \(P_LP_H=0\), and every noise vector can be decomposed as
\[
    z = P_Lz + P_Hz .
\]
For an initial draw \(z^0\), define
\[
    z_L^0=P_Lz^0,\quad
    z_H^0=P_Hz^0,\quad
    r=\|z_L^0\|_2 .
\]

\subsection{Manifold and Tangent Space}

MoNO constrains the optimized noise to preserve both the Gaussian-shell norm and the high-frequency component of the initial noise:
\[
    \mathcal{M}(z^0)
    =
    \left\{
    z\in\mathbb{R}^d:
    \|z\|_2=\|z^0\|_2,\quad
    P_Hz=P_Hz^0
    \right\}.
\]
Because \(P_L\) and \(P_H\) are orthogonal projections, we have
\[
    \|z\|_2^2
    =
    \|P_Lz\|_2^2+\|P_Hz\|_2^2 .
\]
For any \(z\in\mathcal{M}(z^0)\), the high-frequency constraint gives \(P_Hz=z_H^0\). Combining this with the fixed-norm constraint gives
\[
    \|P_Lz\|_2^2
    =
    \|z^0\|_2^2-\|z_H^0\|_2^2
    =
    \|z_L^0\|_2^2
    =
    r^2 .
\]
Therefore, \(\mathcal{M}(z^0)\) can equivalently be written as the affine low-frequency sphere
\[
    \mathcal{M}(z^0)
    =
    \left\{
    z=z_H^0+u:
    u\in\operatorname{Im}(P_L),\quad
    \|u\|_2=r
    \right\}.
\]
The fixed component \(z_H^0\) only translates the sphere, so the nontrivial geometry lies in the low-frequency variable \(u=P_Lz\).

We now derive the tangent space. Let \(z(t)\) be a smooth curve on \(\mathcal{M}(z^0)\) with \(z(0)=z=z_H^0+u\), and let
\[
    \xi=z'(0)
\]
be its velocity. Since \(P_Hz(t)=z_H^0\) for all \(t\), differentiating gives
\[
    P_H\xi=0,
\]
which implies \(\xi\in\operatorname{Im}(P_L)\). Since \(\|P_Lz(t)\|_2^2=r^2\) for all \(t\), differentiating at \(t=0\) gives
\[
    0
    =
    \frac{\mathrm d}{\mathrm dt}\|P_Lz(t)\|_2^2\bigg|_{t=0}
    =
    2\langle P_Lz,\xi\rangle
    =
    2\langle u,\xi\rangle .
\]
Thus, any feasible infinitesimal update must lie in the low-frequency subspace and be orthogonal to the current radius \(u\). The tangent space is therefore
\[
    T_z\mathcal{M}(z^0)
    =
    \left\{
    \xi\in\operatorname{Im}(P_L):
    \langle \xi,u\rangle=0
    \right\},
    \quad
    u=P_Lz .
\]

\newcommand{\proxyvisimg}[1]{%
  \includegraphics[width=0.122\textwidth]{#1}%
}

\begin{figure*}[t]
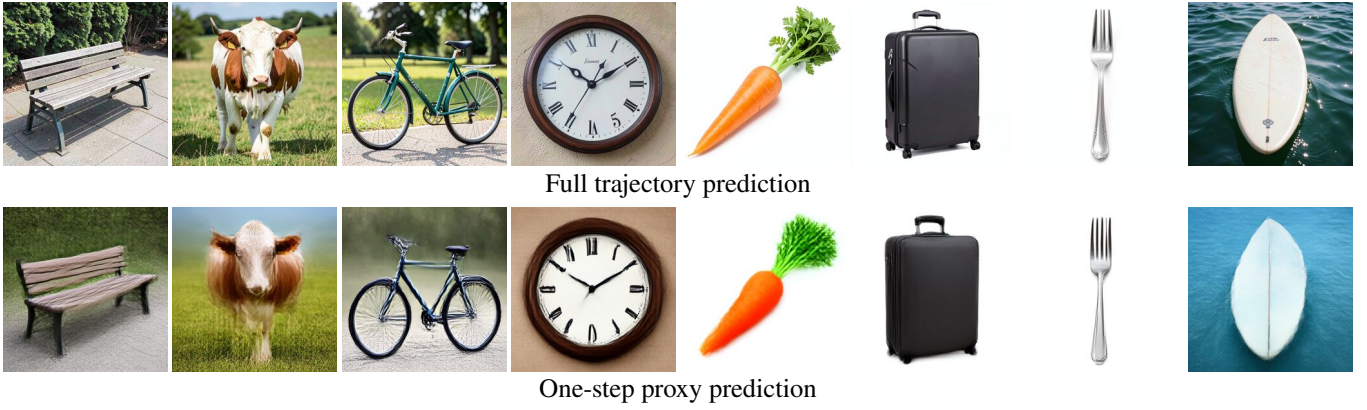

  \centering
  \setlength{\tabcolsep}{1pt}
  \renewcommand{\arraystretch}{0.88}
  \makebox[\textwidth][c]{%
  \begin{tabular}{@{}cccccccc@{}}
    \proxyvisimg{appendix/proxy/full/1.jpg}
    & \proxyvisimg{appendix/proxy/full/2.jpg}
    & \proxyvisimg{appendix/proxy/full/3.jpg}
    & \proxyvisimg{appendix/proxy/full/4.jpg}
    & \proxyvisimg{appendix/proxy/full/5.jpg}
    & \proxyvisimg{appendix/proxy/full/6.jpg}
    & \proxyvisimg{appendix/proxy/full/7.jpg}
    & \proxyvisimg{appendix/proxy/full/8.jpg} \\
    \multicolumn{8}{c}{Full trajectory prediction} \\[3pt]
    \proxyvisimg{appendix/proxy/proxy/1.jpg}
    & \proxyvisimg{appendix/proxy/proxy/2.jpg}
    & \proxyvisimg{appendix/proxy/proxy/3.jpg}
    & \proxyvisimg{appendix/proxy/proxy/4.jpg}
    & \proxyvisimg{appendix/proxy/proxy/5.jpg}
    & \proxyvisimg{appendix/proxy/proxy/6.jpg}
    & \proxyvisimg{appendix/proxy/proxy/7.jpg}
    & \proxyvisimg{appendix/proxy/proxy/8.jpg} \\
    \multicolumn{8}{c}{One-step proxy prediction}
  \end{tabular}}
  \caption{
  Visualization of the proxy predictor using FLUX.2~[klein] 4B.
  The first row shows images generated by the full sampling trajectory, while the
  second row shows images decoded from the corresponding one-step proxy prediction
  under the same seed, prompt, and sampling configuration.
  Although the proxy prediction has lower visual quality than the full-trajectory
  output, it already captures the coarse visual structures needed for diversity
  comparison during optimization.
  }
  \label{fig:app_proxy_predictor}
\end{figure*}

\subsection{Riemannian Gradient}

Let \(f(z)\) denote the diversity objective at the current sequential step, and let
\[
    g=\nabla_z f(z)
\]
be its Euclidean gradient in the ambient noise space. Under the Euclidean metric, the Riemannian gradient is the unique tangent vector \(\operatorname{grad} f(z)\in T_z\mathcal{M}(z^0)\) satisfying
\[
    \langle \operatorname{grad} f(z),\xi\rangle
    =
    \langle g,\xi\rangle
    \quad
    \text{for all }
    \xi\in T_z\mathcal{M}(z^0).
\]
Equivalently, it is the orthogonal projection of the ambient gradient \(g\) onto the tangent space.

The tangent space imposes two constraints. First, feasible directions must lie in \(\operatorname{Im}(P_L)\), so the high-frequency component of \(g\) is removed by \(P_Lg\). Second, feasible directions must be orthogonal to \(u=P_Lz\), so the radial component inside the low-frequency subspace must also be removed. Since \(u\in\operatorname{Im}(P_L)\), the orthogonal projection of \(P_Lg\) onto the complement of \(u\) is
\[
    P_Lg
    -
    \frac{\langle P_Lg,u\rangle}{\|u\|_2^2}u .
\]
Therefore,
\[
    \operatorname{grad} f(z)
    =
    \operatorname{Proj}_{z}^{\mathcal{M}(z^0)}(g)
    =
    P_Lg
    -
    \frac{\langle P_Lg,u\rangle}{\|u\|_2^2}u .
\]
This expression is tangent because
\[
    P_H\operatorname{grad} f(z)=0
\]
and
\[
    \left\langle
    \operatorname{grad} f(z),u
    \right\rangle
    =
    \langle P_Lg,u\rangle
    -
    \frac{\langle P_Lg,u\rangle}{\|u\|_2^2}
    \langle u,u\rangle
    =
    0 .
\]
It is also the closest tangent vector to the ambient gradient in Euclidean norm, since it removes exactly the orthogonal complement of the tangent space: the high-frequency component and the low-frequency radial component.

\subsection{Exponential Map and Constraint Preservation}

We next derive the closed-form exponential-map update. Since \(z_H^0\) is fixed, it suffices to consider the radius-\(r\) sphere in the low-frequency subspace:
\[
    \mathbb{S}_r
    =
    \left\{
    u\in\operatorname{Im}(P_L):
    \|u\|_2=r
    \right\}.
\]
For a tangent vector \(\eta\in T_u\mathbb{S}_r\), we have \(\langle \eta,u\rangle=0\). Consider the curve
\[
    \gamma(t)
    =
    u\cos\left(\frac{t\|\eta\|_2}{r}\right)
    +
    r\frac{\eta}{\|\eta\|_2}
    \sin\left(\frac{t\|\eta\|_2}{r}\right).
\]
This curve starts from \(u\), since
\[
    \gamma(0)=u.
\]
Its initial velocity is
\[
    \gamma'(0)=\eta.
\]
Moreover, because \(u\perp \eta\), its norm is preserved:
\[
    \|\gamma(t)\|_2^2
    =
    \|u\|_2^2
    \cos^2\left(\frac{t\|\eta\|_2}{r}\right)
    +
    r^2
    \sin^2\left(\frac{t\|\eta\|_2}{r}\right)
    =
    r^2 .
\]
Thus \(\gamma(t)\) remains on \(\mathbb{S}_r\). Its acceleration is normal to the sphere, so \(\gamma(t)\) is the geodesic on the radius-\(r\) sphere with initial point \(u\) and initial velocity \(\eta\). Hence the exponential map is
\[
    \operatorname{Exp}_{u}^{\mathbb{S}_r}(\eta)
    =
    u\cos\left(\frac{\|\eta\|_2}{r}\right)
    +
    r\frac{\eta}{\|\eta\|_2}
    \sin\left(\frac{\|\eta\|_2}{r}\right),
\]
with the convention \(\operatorname{Exp}_{u}^{\mathbb{S}_r}(0)=u\).

At iteration \(k\), MoNO takes the Riemannian descent direction
\[
    h^k=-\operatorname{grad} f(z^k).
\]
Since \(h^k\in T_{z^k}\mathcal{M}(z^0)\), it is also a tangent vector for the low-frequency sphere at \(u^k=P_Lz^k\). Setting \(\eta=\alpha h^k\), the exponential-map update gives
\[
    u^{k+1}
    =
    u^k
    \cos\left(\frac{\alpha\|h^k\|_2}{r}\right)
    +
    r\frac{h^k}{\|h^k\|_2}
    \sin\left(\frac{\alpha\|h^k\|_2}{r}\right).
\]
The full noise is then reconstructed by adding back the fixed high-frequency component:
\[
    z^{k+1}=z_H^0+u^{k+1}.
\]
Equivalently,
\[
    z^{k+1}
    =
    \operatorname{Exp}_{z^k}^{\mathcal{M}(z^0)}
    \left(
    \alpha h^k
    \right).
\]

Finally, the update preserves the constraints exactly. Since \(u^{k+1}\in\operatorname{Im}(P_L)\),
\[
    P_Hz^{k+1}
    =
    P_H(z_H^0+u^{k+1})
    =
    z_H^0 .
\]
Since the exponential map remains on the radius-\(r\) sphere,
\[
    \|P_Lz^{k+1}\|_2
    =
    \|u^{k+1}\|_2
    =
    r
    =
    \|P_Lz^0\|_2 .
\]
Therefore,
\[
    z^{k+1}\in\mathcal{M}(z^0),
\]
so every Riemannian update step remains feasible by construction.

\begin{table*}[t]
\centering
\setlength{\tabcolsep}{4pt}    
\renewcommand{\arraystretch}{1.12}
\caption{
Results on DPG-Bench with distilled text-to-image diffusion models.
We report the mean across prompts, with the standard deviation shown in subscript.
The best results within each model block are shown in bold.
}
\label{tab:main_dpg}
\begin{tabular}{c|l |ccc | ccc}
\toprule
\multirow{2}{*}{Model}
& \multirow{2}{*}{Method}
& \multicolumn{3}{c|}{Diversity $\uparrow$}
& \multicolumn{3}{c}{Quality $\uparrow$} \\
\cmidrule(lr){3-5}
\cmidrule(lr){6-8}
& & LPIPS & DreamSim & Vendi Score
& CLIPScore & CLIP-IQA & HPSv2 \\
\midrule
\multirow{9}{*}{\shortstack[c]{FLUX.2 \\[3pt] {[klein]} 4B \\[3pt] (4 steps)}}
& i.i.d.
& \ms{0.543}{0.102} & \ms{0.121}{0.045} & \ms{1.589}{0.292}
& \ms{0.342}{0.035} & \ms{0.678}{0.125} & \ms{0.293}{0.035} \\
& CADS
& \ms{0.698}{0.048} & \ms{0.275}{0.073} & \ms{2.087}{0.444}
& \ms{0.339}{0.033} & \ms{0.626}{0.130} & \ms{0.282}{0.033} \\
& Particle Guidance
& \ms{0.543}{0.103} & \ms{0.121}{0.045} & \ms{1.588}{0.298}
& \ms{0.343}{0.035} & \ms{0.681}{0.126} & \ms{0.294}{0.035} \\
& NegToMe
& \ms{0.550}{0.103} & \ms{0.124}{0.046} & \ms{1.602}{0.303}
& \ms{0.343}{0.035} & \ms{0.674}{0.125} & \ms{0.293}{0.035} \\
& Group Inference
& \ms{0.588}{0.094} & \ms{0.144}{0.051} & \ms{1.673}{0.328}
& \bms{0.344}{0.035} & \ms{0.678}{0.125} & \ms{0.293}{0.035} \\
& DivGen
& \ms{0.550}{0.099} & \ms{0.124}{0.046} & \ms{1.621}{0.312}
& \bms{0.344}{0.035} & \bms{0.682}{0.126} & \bms{0.296}{0.035} \\
& CNO
& \ms{0.543}{0.103} & \ms{0.121}{0.045} & \ms{1.592}{0.299}
& \ms{0.343}{0.035} & \ms{0.681}{0.126} & \ms{0.294}{0.035} \\
\rowcolor{oursgray}
\cellcolor{white} & MoNO (1-iter)
& \ms{0.665}{0.075} & \ms{0.218}{0.067} & \ms{1.880}{0.377}
& \ms{0.341}{0.034} & \ms{0.670}{0.122} & \ms{0.288}{0.033} \\
\rowcolor{oursgray}
\cellcolor{white} & MoNO (4-iter)
& \bms{0.714}{0.066} & \bms{0.301}{0.079} & \bms{2.151}{0.452}
& \ms{0.338}{0.033} & \ms{0.661}{0.120} & \ms{0.280}{0.031} \\
\midrule
& i.i.d.
& \ms{0.623}{0.059} & \ms{0.176}{0.063} & \ms{1.787}{0.375}
& \bms{0.343}{0.033} & \ms{0.641}{0.131} & \bms{0.284}{0.037} \\
& DivGen
& \ms{0.639}{0.061} & \ms{0.207}{0.072} & \ms{2.027}{0.445}
& \ms{0.341}{0.032} & \ms{0.620}{0.132} & \ms{0.281}{0.037} \\
& CNO
& \ms{0.633}{0.063} & \ms{0.193}{0.067} & \ms{1.829}{0.391}
& \ms{0.342}{0.032} & \ms{0.620}{0.132} & \ms{0.282}{0.037} \\
\rowcolor{oursgray}
\cellcolor{white} & MoNO (1-iter)
& \ms{0.674}{0.054} & \ms{0.253}{0.080} & \ms{2.001}{0.433}
& \ms{0.340}{0.032} & \ms{0.643}{0.126} & \bms{0.284}{0.035} \\
\rowcolor{oursgray}
\cellcolor{white}\multirow{-5}{*}{\shortstack[c]{SDXL-Turbo \\[3pt] (1 step)}} & MoNO (4-iter)
& \bms{0.711}{0.055} & \bms{0.315}{0.091} & \bms{2.225}{0.491}
& \ms{0.338}{0.031} & \bms{0.648}{0.119} & \ms{0.281}{0.033} \\
\midrule
& i.i.d.
& \ms{0.634}{0.081} & \ms{0.169}{0.055} & \ms{1.866}{0.384}
& \bms{0.347}{0.032} & \ms{0.668}{0.110} & \ms{0.279}{0.037} \\
& DivGen
& \ms{0.652}{0.078} & \ms{0.190}{0.063} & \ms{2.083}{0.441}
& \ms{0.346}{0.032} & \ms{0.659}{0.113} & \bms{0.281}{0.037} \\
& CNO
& \ms{0.644}{0.081} & \ms{0.177}{0.058} & \ms{1.893}{0.383}
& \bms{0.347}{0.032} & \ms{0.658}{0.113} & \bms{0.281}{0.037} \\
\rowcolor{oursgray}
\cellcolor{white} & MoNO (1-iter)
& \ms{0.707}{0.070} & \ms{0.253}{0.071} & \ms{2.099}{0.445}
& \ms{0.345}{0.031} & \ms{0.678}{0.102} & \ms{0.280}{0.034} \\
\rowcolor{oursgray}
\cellcolor{white}\multirow{-5}{*}{\shortstack[c]{DMD2 \\[3pt] (1 step)}} & MoNO (4-iter)
& \bms{0.741}{0.065} & \bms{0.296}{0.074} & \bms{2.268}{0.478}
& \ms{0.342}{0.031} & \bms{0.684}{0.097} & \ms{0.279}{0.033} \\
\bottomrule
\end{tabular}
\end{table*}

\section{Detailed Experimental Settings}

\subsection{General Setup}

All experiments are conducted on NVIDIA A100-SXM4-40GB hardware. 
DMD2 is evaluated at a resolution of $1024 \times 1024$, while the other generators are evaluated at $512 \times 512$. 
For MoNO, we use \texttt{facebook/dinov2-base}~\citep{oquab2023dinov2} as the frozen feature extractor, and enable gradient checkpointing during optimization to reduce memory usage. 
All generated images are evaluated using the same metric implementation. 
Specifically, we report LPIPS~\citep{zhang2018unreasonable} with a VGG backbone, DreamSim~\citep{fu2023dreamsim} with the \texttt{dino\_vitb16} backbone, and Vendi Score~\citep{friedman2022vendi} computed from \texttt{facebook/dinov2-base}~\citep{oquab2023dinov2} features as diversity metrics. 
For quality evaluation, we report CLIPScore~\citep{hessel2021clipscore} using \texttt{openai/clip-vit-base-patch32}~\citep{radford2021learning}, CLIP-IQA~\citep{wang2023exploring} using the \texttt{pyiqa} implementation with the \texttt{clipiqa} model, and HPSv2~\citep{wu2023human} using the \texttt{xswu/HPSv2} checkpoint.

\begin{figure*}[t]
  \centering
  \setlength{\tabcolsep}{2pt}
  \begin{tabular}{@{}ccc@{}}
    \includegraphics[width=0.325\textwidth]{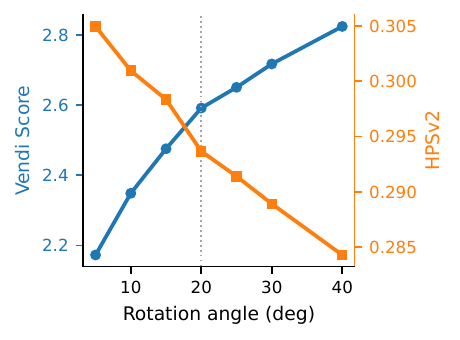}
    &
    \includegraphics[width=0.325\textwidth]{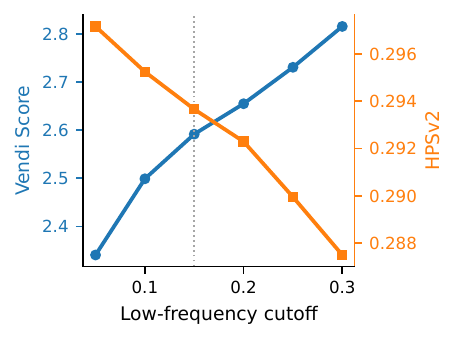}
    &
    \includegraphics[width=0.325\textwidth]{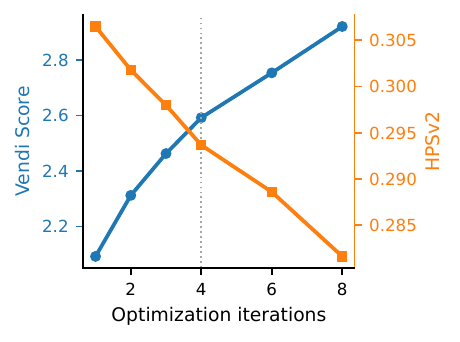}
  \end{tabular}
  \caption{
  Hyperparameter sweeps of MoNO over rotation angle, low-frequency cutoff, and optimization iterations.
  Each plot uses Vendi Score as the diversity metric and HPSv2 as the quality metric.
  }
  \label{fig:app_sweep_tradeoff}
\end{figure*}

\subsection{MoNO Implementation Details}
MoNO generates images sequentially for each prompt. 
The first image is sampled without optimization because the feature basis is empty. Each subsequent image is optimized against the patch-wise feature subspace constructed from previously generated images.

For the low-frequency constraint, we implement \(P_L\) by applying a spatial Fourier low-pass projection independently to each latent channel. 
Let \(F\) denote the two-dimensional discrete Fourier transform over spatial dimensions, and let \(M_{\rho}\) be a binary circular low-pass mask. 
The low- and high-frequency projections are defined as
\[
    P_L 
    =
    F^{-1}
    M_{\rho}
    F ,
    \quad
    P_H 
    =
    I - P_L  .
\]
The mask is given by
\[
    M_{\rho}(\omega_x,\omega_y)
    =
    \mathbf{1}
    \left[
    \frac{\sqrt{\omega_x^2+\omega_y^2}}{\sqrt{1/2}}
    \leq \rho
    \right],
\]
where \(\omega_x\) and \(\omega_y\) are normalized Fourier frequencies. 
We use a cutoff ratio of \(\rho=0.15\) in all experiments. 
This gives the constrained search space
\[
    \mathcal{M}
    =
    \left\{
    z=P_L z + P_H z^0
    :
    \|P_L z\|_2 = \|P_L z^0\|_2
    \right\},
\]
where the high-frequency component is fixed to that of the initial noise \(z^0\), and only the low-frequency component is rotated on a sphere.

For the geodesic update, we tune the induced rotation angle rather than the raw Euclidean step size. 
Let \(u^k=P_Lz^k\), \(r=\|P_Lz^0\|_2\), and \(h^k\) be the projected descent direction on the low-frequency tangent space. 
The angular step is parameterized as
\[
\begin{aligned}
    \varphi^k
    &=
    \frac{\alpha\|h^k\|_2}{r}.
\end{aligned}
\]
The low-frequency component is then updated by
\[
\begin{aligned}
    u^{k+1}
    &=
    u^k\cos\varphi^k
    +
    r
    \frac{h^k}{\|h^k\|_2}
    \sin\varphi^k, \\
    z^{k+1}
    &=
    P_Hz^0 + u^{k+1}.
\end{aligned}
\]
Since the sphere has an explicit geometric structure, using the rotation angle as the hyperparameter makes the step size easier to tune and transfer across settings. 
We set \(\varphi=30^\circ\) for the one-iteration variant and \(\varphi=20^\circ\) for the four-iteration variant.

For efficient optimization, we use proxy predictor \(A_\theta\) to obtain a clean-image estimate from the current initial noise for feature extraction.
For one-step generators, this estimate is simply given by the one-step clean prediction used by the sampler itself.
For multi-step generators like FLUX.2~[klein] 4B, we instead use the model's one-step prediction as a proxy, avoiding backpropagation through the full sampling trajectory.
Specifically, let \(D_{\mathrm{VAE}}\) denote the VAE decoder, and let \(T\) denote the initial denoising timestep.
For FLUX.2~[klein] 4B, a flow-matching model, the network predicts a velocity field \(v_\theta(z_T,T,c)\)\footnote{Here the subscript \(T\) denotes the denoising timestep, not the sequential generation index.}. 
Under the scheduler's linear interpolation convention
\(
    z_T = (1-\sigma_T)z_0 + \sigma_T\epsilon,
\)
we use the corresponding first-order Euler endpoint estimate
\[
    \hat{z}_{0}
    =
    z_T-\sigma_T v_\theta(z_T,T,c),
    \quad
    A_\theta(z_T,c)
    =
    D_{\mathrm{VAE}}\!\left(\hat{z}_{0}\right),
\]
where \(\sigma_T\) denotes the initial flow noise level under the sampler parameterization. 
The estimate \(A_\theta\) is used only for feature extraction and optimization.
After optimization, the final image is still generated by running the original sampler \(G_\theta(z,c)\) from the optimized initial noise.
\Cref{fig:app_proxy_predictor} visualizes this proxy on FLUX.2~[klein] 4B by comparing images generated from the full sampling trajectory with images decoded from the corresponding proxy prediction under the same sampling settings, showing that the proxy already captures the coarse visual structure needed for diversity comparison during optimization.

For the diversity objective, we extract DINOv2 patch features from the predicted clean image used for optimization. 
Specifically, the predicted image is resized to \(256\times256\), the CLS token is discarded, and every patch feature is \(\ell_2\)-normalized. 
Let \(\psi_p(\hat{x})\in\mathbb{R}^D\) denote the normalized DINO feature of patch \(p\), and let \(B_{p,n-1}\in\mathbb{R}^{D\times m_{p,n-1}}\) be the orthonormal basis maintained at the same patch position from the previous \(n-1\) generated images, where \(m_{p,n-1}\) denotes the current basis size.
For the \(n\)-th image, the optimized diversity loss is the average patch-wise projection length,
\[
    \mathcal{L}_{\mathrm{div}}(z_n)
    =
    \frac{1}{P}
    \sum_{p=1}^{P}
    \left\|
    B_{p,n-1}^{\top}
    \psi_p\!\left(A_{\theta}(z_n,c)\right)
    \right\|_2^2 .
\]
After the final image \(G_\theta(z_n,c)\) is generated, its DINO patch features are added to the maintained patch-wise bases using Gram--Schmidt orthogonalization.

\subsection{Compared Method Details}
For Group Inference~\citep{parmar2025scaling}, we follow its default setting, starting from \(64\) candidates for each prompt and progressively pruning them using the original group-level objective with CLIP~\citep{radford2021learning} text-image similarity as the unary term and DINO~\citep{oquab2023dinov2} feature diversity as the binary term.
Other selection hyperparameters follow the original default settings.
For DivGen~\citep{harrington2026s}, we keep its original reward-based quality regularization, including HPS-based quality control and CLIP regularization when used by the corresponding configuration, and use the DPP-based diversity objective with DINO features.
Since the official implementations of CADS~\citep{sadat2024cads} and CNO~\citep{kim2025diverse} are not publicly available at the time of our experiments, we re-implement both methods according to the algorithmic descriptions in their papers.
For CADS, we perturb the text conditioning with scheduled Gaussian noise following the condition-annealing schedule.
For CNO, we follow its contrastive noise optimization objective and preserve the anchoring regularization described in the original method.
All baseline-specific hyperparameters are fixed across prompts and datasets.

\section{Additional Experimental Results}

\subsection{Additional Quantitative Results}
We further evaluate MoNO on DPG-Bench~\citep{hu2024ella} to verify its effectiveness beyond the main evaluation setting.
Compared with GenEval~\citep{ghosh2023geneval}, which focuses on structured compositional prompts with controlled object-level attributes and relations, DPG-Bench provides a complementary testbed with denser prompts and richer semantic details.
As shown in \Cref{tab:main_dpg}, MoNO consistently improves diversity across the evaluated models.
In particular, the four-iteration variant achieves the best diversity scores within each model block, while maintaining competitive image quality.
These results further support that constrained noise optimization can improve sample diversity without relying on external reward models or large candidate pools.

\subsection{Hyperparameter Sensitivity}

We analyze the sensitivity of MoNO to three key hyperparameters: the rotation angle \(\varphi\),
the low-frequency cutoff \(\rho\), and the number of optimization iterations \(K\).
All sweeps are conducted on GenEval using FLUX.2~[klein] 4B, around the default configuration of four optimization iterations,
a rotation angle of \(20^\circ\), and a low-frequency cutoff of \(0.15\).
For each sweep, we vary only one hyperparameter while keeping the other two fixed.
As shown in \Cref{fig:app_sweep_tradeoff}, changing these hyperparameters induces a clear
quality--diversity trade-off: stronger optimization improves diversity, as reflected by
higher Vendi Score, while slightly reducing image quality, as reflected by lower HPSv2.
These results show that MoNO provides a simple and controllable way to balance diversity
and quality.

\subsection{Additional Qualitative Visualization}
We provide additional qualitative comparisons between i.i.d.\ sampling and MoNO sampling for three distilled diffusion models.
For each prompt, we visualize \(16\) images each from i.i.d.\ sampling and MoNO sampling.
We show the one-iteration variant of MoNO in \Cref{fig:app_vis_flux_klein_1iter,fig:app_vis_sdxl_turbo_1iter,fig:app_vis_dmd2_1iter}, and the four-iteration variant in \Cref{fig:app_vis_flux_klein_4iter,fig:app_vis_sdxl_turbo_4iter,fig:app_vis_dmd2_4iter}.
Across different models, prompts, and optimization budgets, MoNO produces visibly more diverse samples than i.i.d.\ sampling while preserving generation fidelity.

\newcommand{\appvisimg}[1]{%
  \includegraphics[height=0.104\textheight]{#1}%
}

\begin{figure*}[p]
  \centering
  \setlength{\tabcolsep}{1pt}
  \renewcommand{\arraystretch}{0.88}
  \makebox[\textwidth][c]{%
  \begin{tabular}{@{}cccc@{}}
    \appvisimg{appendix/visualization/1iter/flux_klein/iid/1.jpg}
    & \appvisimg{appendix/visualization/1iter/flux_klein/iid/2.jpg}
    & \appvisimg{appendix/visualization/1iter/flux_klein/iid/3.jpg}
    & \appvisimg{appendix/visualization/1iter/flux_klein/iid/4.jpg} \\[-2pt]
    \appvisimg{appendix/visualization/1iter/flux_klein/iid/5.jpg}
    & \appvisimg{appendix/visualization/1iter/flux_klein/iid/6.jpg}
    & \appvisimg{appendix/visualization/1iter/flux_klein/iid/7.jpg}
    & \appvisimg{appendix/visualization/1iter/flux_klein/iid/8.jpg} \\[-2pt]
    \appvisimg{appendix/visualization/1iter/flux_klein/iid/9.jpg}
    & \appvisimg{appendix/visualization/1iter/flux_klein/iid/10.jpg}
    & \appvisimg{appendix/visualization/1iter/flux_klein/iid/11.jpg}
    & \appvisimg{appendix/visualization/1iter/flux_klein/iid/12.jpg} \\[-2pt]
    \appvisimg{appendix/visualization/1iter/flux_klein/iid/13.jpg}
    & \appvisimg{appendix/visualization/1iter/flux_klein/iid/14.jpg}
    & \appvisimg{appendix/visualization/1iter/flux_klein/iid/15.jpg}
    & \appvisimg{appendix/visualization/1iter/flux_klein/iid/16.jpg} \\
    \multicolumn{4}{c}{i.i.d.\ sampling} \\[4pt]
    \appvisimg{appendix/visualization/1iter/flux_klein/mono/1.jpg}
    & \appvisimg{appendix/visualization/1iter/flux_klein/mono/2.jpg}
    & \appvisimg{appendix/visualization/1iter/flux_klein/mono/3.jpg}
    & \appvisimg{appendix/visualization/1iter/flux_klein/mono/4.jpg} \\[-2pt]
    \appvisimg{appendix/visualization/1iter/flux_klein/mono/5.jpg}
    & \appvisimg{appendix/visualization/1iter/flux_klein/mono/6.jpg}
    & \appvisimg{appendix/visualization/1iter/flux_klein/mono/7.jpg}
    & \appvisimg{appendix/visualization/1iter/flux_klein/mono/8.jpg} \\[-2pt]
    \appvisimg{appendix/visualization/1iter/flux_klein/mono/9.jpg}
    & \appvisimg{appendix/visualization/1iter/flux_klein/mono/10.jpg}
    & \appvisimg{appendix/visualization/1iter/flux_klein/mono/11.jpg}
    & \appvisimg{appendix/visualization/1iter/flux_klein/mono/12.jpg} \\[-2pt]
    \appvisimg{appendix/visualization/1iter/flux_klein/mono/13.jpg}
    & \appvisimg{appendix/visualization/1iter/flux_klein/mono/14.jpg}
    & \appvisimg{appendix/visualization/1iter/flux_klein/mono/15.jpg}
    & \appvisimg{appendix/visualization/1iter/flux_klein/mono/16.jpg} \\
    \multicolumn{4}{c}{MoNO sampling}
  \end{tabular}}
  \caption{
  Qualitative comparison between i.i.d.\ sampling and MoNO (1-iter) using FLUX.2~[klein] 4B for the prompt \textit{``a photo of a bench''}.
  }
  \label{fig:app_vis_flux_klein_1iter}
\end{figure*}

\begin{figure*}[p]
  \centering
  \setlength{\tabcolsep}{1pt}
  \renewcommand{\arraystretch}{0.88}
  \makebox[\textwidth][c]{%
  \begin{tabular}{@{}cccc@{}}
    \appvisimg{appendix/visualization/1iter/sdxl_turbo/iid/1.jpg}
    & \appvisimg{appendix/visualization/1iter/sdxl_turbo/iid/2.jpg}
    & \appvisimg{appendix/visualization/1iter/sdxl_turbo/iid/3.jpg}
    & \appvisimg{appendix/visualization/1iter/sdxl_turbo/iid/4.jpg} \\[-2pt]
    \appvisimg{appendix/visualization/1iter/sdxl_turbo/iid/5.jpg}
    & \appvisimg{appendix/visualization/1iter/sdxl_turbo/iid/6.jpg}
    & \appvisimg{appendix/visualization/1iter/sdxl_turbo/iid/7.jpg}
    & \appvisimg{appendix/visualization/1iter/sdxl_turbo/iid/8.jpg} \\[-2pt]
    \appvisimg{appendix/visualization/1iter/sdxl_turbo/iid/9.jpg}
    & \appvisimg{appendix/visualization/1iter/sdxl_turbo/iid/10.jpg}
    & \appvisimg{appendix/visualization/1iter/sdxl_turbo/iid/11.jpg}
    & \appvisimg{appendix/visualization/1iter/sdxl_turbo/iid/12.jpg} \\[-2pt]
    \appvisimg{appendix/visualization/1iter/sdxl_turbo/iid/13.jpg}
    & \appvisimg{appendix/visualization/1iter/sdxl_turbo/iid/14.jpg}
    & \appvisimg{appendix/visualization/1iter/sdxl_turbo/iid/15.jpg}
    & \appvisimg{appendix/visualization/1iter/sdxl_turbo/iid/16.jpg} \\
    \multicolumn{4}{c}{i.i.d.\ sampling} \\[4pt]
    \appvisimg{appendix/visualization/1iter/sdxl_turbo/mono/1.jpg}
    & \appvisimg{appendix/visualization/1iter/sdxl_turbo/mono/2.jpg}
    & \appvisimg{appendix/visualization/1iter/sdxl_turbo/mono/3.jpg}
    & \appvisimg{appendix/visualization/1iter/sdxl_turbo/mono/4.jpg} \\[-2pt]
    \appvisimg{appendix/visualization/1iter/sdxl_turbo/mono/5.jpg}
    & \appvisimg{appendix/visualization/1iter/sdxl_turbo/mono/6.jpg}
    & \appvisimg{appendix/visualization/1iter/sdxl_turbo/mono/7.jpg}
    & \appvisimg{appendix/visualization/1iter/sdxl_turbo/mono/8.jpg} \\[-2pt]
    \appvisimg{appendix/visualization/1iter/sdxl_turbo/mono/9.jpg}
    & \appvisimg{appendix/visualization/1iter/sdxl_turbo/mono/10.jpg}
    & \appvisimg{appendix/visualization/1iter/sdxl_turbo/mono/11.jpg}
    & \appvisimg{appendix/visualization/1iter/sdxl_turbo/mono/12.jpg} \\[-2pt]
    \appvisimg{appendix/visualization/1iter/sdxl_turbo/mono/13.jpg}
    & \appvisimg{appendix/visualization/1iter/sdxl_turbo/mono/14.jpg}
    & \appvisimg{appendix/visualization/1iter/sdxl_turbo/mono/15.jpg}
    & \appvisimg{appendix/visualization/1iter/sdxl_turbo/mono/16.jpg} \\
    \multicolumn{4}{c}{MoNO sampling}
  \end{tabular}}
  \caption{
  Qualitative comparison between i.i.d.\ sampling and MoNO (1-iter) using SDXL-Turbo for the prompt \textit{``a photo of a car''}.
  }
  \label{fig:app_vis_sdxl_turbo_1iter}
\end{figure*}

\begin{figure*}[p]
  \centering
  \setlength{\tabcolsep}{1pt}
  \renewcommand{\arraystretch}{0.88}
  \makebox[\textwidth][c]{%
  \begin{tabular}{@{}cccc@{}}
    \appvisimg{appendix/visualization/1iter/dmd2/iid/1.jpg}
    & \appvisimg{appendix/visualization/1iter/dmd2/iid/2.jpg}
    & \appvisimg{appendix/visualization/1iter/dmd2/iid/3.jpg}
    & \appvisimg{appendix/visualization/1iter/dmd2/iid/4.jpg} \\[-2pt]
    \appvisimg{appendix/visualization/1iter/dmd2/iid/5.jpg}
    & \appvisimg{appendix/visualization/1iter/dmd2/iid/6.jpg}
    & \appvisimg{appendix/visualization/1iter/dmd2/iid/7.jpg}
    & \appvisimg{appendix/visualization/1iter/dmd2/iid/8.jpg} \\[-2pt]
    \appvisimg{appendix/visualization/1iter/dmd2/iid/9.jpg}
    & \appvisimg{appendix/visualization/1iter/dmd2/iid/10.jpg}
    & \appvisimg{appendix/visualization/1iter/dmd2/iid/11.jpg}
    & \appvisimg{appendix/visualization/1iter/dmd2/iid/12.jpg} \\[-2pt]
    \appvisimg{appendix/visualization/1iter/dmd2/iid/13.jpg}
    & \appvisimg{appendix/visualization/1iter/dmd2/iid/14.jpg}
    & \appvisimg{appendix/visualization/1iter/dmd2/iid/15.jpg}
    & \appvisimg{appendix/visualization/1iter/dmd2/iid/16.jpg} \\
    \multicolumn{4}{c}{i.i.d.\ sampling} \\[4pt]
    \appvisimg{appendix/visualization/1iter/dmd2/mono/1.jpg}
    & \appvisimg{appendix/visualization/1iter/dmd2/mono/2.jpg}
    & \appvisimg{appendix/visualization/1iter/dmd2/mono/3.jpg}
    & \appvisimg{appendix/visualization/1iter/dmd2/mono/4.jpg} \\[-2pt]
    \appvisimg{appendix/visualization/1iter/dmd2/mono/5.jpg}
    & \appvisimg{appendix/visualization/1iter/dmd2/mono/6.jpg}
    & \appvisimg{appendix/visualization/1iter/dmd2/mono/7.jpg}
    & \appvisimg{appendix/visualization/1iter/dmd2/mono/8.jpg} \\[-2pt]
    \appvisimg{appendix/visualization/1iter/dmd2/mono/9.jpg}
    & \appvisimg{appendix/visualization/1iter/dmd2/mono/10.jpg}
    & \appvisimg{appendix/visualization/1iter/dmd2/mono/11.jpg}
    & \appvisimg{appendix/visualization/1iter/dmd2/mono/12.jpg} \\[-2pt]
    \appvisimg{appendix/visualization/1iter/dmd2/mono/13.jpg}
    & \appvisimg{appendix/visualization/1iter/dmd2/mono/14.jpg}
    & \appvisimg{appendix/visualization/1iter/dmd2/mono/15.jpg}
    & \appvisimg{appendix/visualization/1iter/dmd2/mono/16.jpg} \\
    \multicolumn{4}{c}{MoNO sampling}
  \end{tabular}}
  \caption{
  Qualitative comparison between i.i.d.\ sampling and MoNO (1-iter) using DMD2 for the prompt \textit{``a photo of a bed''}.
  }
  \label{fig:app_vis_dmd2_1iter}
\end{figure*}

\begin{figure*}[p]
  \centering
  \setlength{\tabcolsep}{1pt}
  \renewcommand{\arraystretch}{0.88}
  \makebox[\textwidth][c]{%
  \begin{tabular}{@{}cccc@{}}
    \appvisimg{appendix/visualization/4iter/flux_klein/iid/1.jpg}
    & \appvisimg{appendix/visualization/4iter/flux_klein/iid/2.jpg}
    & \appvisimg{appendix/visualization/4iter/flux_klein/iid/3.jpg}
    & \appvisimg{appendix/visualization/4iter/flux_klein/iid/4.jpg} \\[-2pt]
    \appvisimg{appendix/visualization/4iter/flux_klein/iid/5.jpg}
    & \appvisimg{appendix/visualization/4iter/flux_klein/iid/6.jpg}
    & \appvisimg{appendix/visualization/4iter/flux_klein/iid/7.jpg}
    & \appvisimg{appendix/visualization/4iter/flux_klein/iid/8.jpg} \\[-2pt]
    \appvisimg{appendix/visualization/4iter/flux_klein/iid/9.jpg}
    & \appvisimg{appendix/visualization/4iter/flux_klein/iid/10.jpg}
    & \appvisimg{appendix/visualization/4iter/flux_klein/iid/11.jpg}
    & \appvisimg{appendix/visualization/4iter/flux_klein/iid/12.jpg} \\[-2pt]
    \appvisimg{appendix/visualization/4iter/flux_klein/iid/13.jpg}
    & \appvisimg{appendix/visualization/4iter/flux_klein/iid/14.jpg}
    & \appvisimg{appendix/visualization/4iter/flux_klein/iid/15.jpg}
    & \appvisimg{appendix/visualization/4iter/flux_klein/iid/16.jpg} \\
    \multicolumn{4}{c}{i.i.d.\ sampling} \\[4pt]
    \appvisimg{appendix/visualization/4iter/flux_klein/mono/1.jpg}
    & \appvisimg{appendix/visualization/4iter/flux_klein/mono/2.jpg}
    & \appvisimg{appendix/visualization/4iter/flux_klein/mono/3.jpg}
    & \appvisimg{appendix/visualization/4iter/flux_klein/mono/4.jpg} \\[-2pt]
    \appvisimg{appendix/visualization/4iter/flux_klein/mono/5.jpg}
    & \appvisimg{appendix/visualization/4iter/flux_klein/mono/6.jpg}
    & \appvisimg{appendix/visualization/4iter/flux_klein/mono/7.jpg}
    & \appvisimg{appendix/visualization/4iter/flux_klein/mono/8.jpg} \\[-2pt]
    \appvisimg{appendix/visualization/4iter/flux_klein/mono/9.jpg}
    & \appvisimg{appendix/visualization/4iter/flux_klein/mono/10.jpg}
    & \appvisimg{appendix/visualization/4iter/flux_klein/mono/11.jpg}
    & \appvisimg{appendix/visualization/4iter/flux_klein/mono/12.jpg} \\[-2pt]
    \appvisimg{appendix/visualization/4iter/flux_klein/mono/13.jpg}
    & \appvisimg{appendix/visualization/4iter/flux_klein/mono/14.jpg}
    & \appvisimg{appendix/visualization/4iter/flux_klein/mono/15.jpg}
    & \appvisimg{appendix/visualization/4iter/flux_klein/mono/16.jpg} \\
    \multicolumn{4}{c}{MoNO sampling}
  \end{tabular}}
  \caption{
  Qualitative comparison between i.i.d.\ sampling and MoNO (4-iter) using FLUX.2~[klein] 4B for the prompt \textit{``a photo of a teddy bear''}.
  }
  \label{fig:app_vis_flux_klein_4iter}
\end{figure*}

\begin{figure*}[p]
  \centering
  \setlength{\tabcolsep}{1pt}
  \renewcommand{\arraystretch}{0.88}
  \makebox[\textwidth][c]{%
  \begin{tabular}{@{}cccc@{}}
    \appvisimg{appendix/visualization/4iter/sdxl_turbo/iid/1.jpg}
    & \appvisimg{appendix/visualization/4iter/sdxl_turbo/iid/2.jpg}
    & \appvisimg{appendix/visualization/4iter/sdxl_turbo/iid/3.jpg}
    & \appvisimg{appendix/visualization/4iter/sdxl_turbo/iid/4.jpg} \\[-2pt]
    \appvisimg{appendix/visualization/4iter/sdxl_turbo/iid/5.jpg}
    & \appvisimg{appendix/visualization/4iter/sdxl_turbo/iid/6.jpg}
    & \appvisimg{appendix/visualization/4iter/sdxl_turbo/iid/7.jpg}
    & \appvisimg{appendix/visualization/4iter/sdxl_turbo/iid/8.jpg} \\[-2pt]
    \appvisimg{appendix/visualization/4iter/sdxl_turbo/iid/9.jpg}
    & \appvisimg{appendix/visualization/4iter/sdxl_turbo/iid/10.jpg}
    & \appvisimg{appendix/visualization/4iter/sdxl_turbo/iid/11.jpg}
    & \appvisimg{appendix/visualization/4iter/sdxl_turbo/iid/12.jpg} \\[-2pt]
    \appvisimg{appendix/visualization/4iter/sdxl_turbo/iid/13.jpg}
    & \appvisimg{appendix/visualization/4iter/sdxl_turbo/iid/14.jpg}
    & \appvisimg{appendix/visualization/4iter/sdxl_turbo/iid/15.jpg}
    & \appvisimg{appendix/visualization/4iter/sdxl_turbo/iid/16.jpg} \\
    \multicolumn{4}{c}{i.i.d.\ sampling} \\[4pt]
    \appvisimg{appendix/visualization/4iter/sdxl_turbo/mono/1.jpg}
    & \appvisimg{appendix/visualization/4iter/sdxl_turbo/mono/2.jpg}
    & \appvisimg{appendix/visualization/4iter/sdxl_turbo/mono/3.jpg}
    & \appvisimg{appendix/visualization/4iter/sdxl_turbo/mono/4.jpg} \\[-2pt]
    \appvisimg{appendix/visualization/4iter/sdxl_turbo/mono/5.jpg}
    & \appvisimg{appendix/visualization/4iter/sdxl_turbo/mono/6.jpg}
    & \appvisimg{appendix/visualization/4iter/sdxl_turbo/mono/7.jpg}
    & \appvisimg{appendix/visualization/4iter/sdxl_turbo/mono/8.jpg} \\[-2pt]
    \appvisimg{appendix/visualization/4iter/sdxl_turbo/mono/9.jpg}
    & \appvisimg{appendix/visualization/4iter/sdxl_turbo/mono/10.jpg}
    & \appvisimg{appendix/visualization/4iter/sdxl_turbo/mono/11.jpg}
    & \appvisimg{appendix/visualization/4iter/sdxl_turbo/mono/12.jpg} \\[-2pt]
    \appvisimg{appendix/visualization/4iter/sdxl_turbo/mono/13.jpg}
    & \appvisimg{appendix/visualization/4iter/sdxl_turbo/mono/14.jpg}
    & \appvisimg{appendix/visualization/4iter/sdxl_turbo/mono/15.jpg}
    & \appvisimg{appendix/visualization/4iter/sdxl_turbo/mono/16.jpg} \\
    \multicolumn{4}{c}{MoNO sampling}
  \end{tabular}}
  \caption{
  Qualitative comparison between i.i.d.\ sampling and MoNO (4-iter) using SDXL-Turbo for the prompt \textit{``a photo of a chair''}.
  }
  \label{fig:app_vis_sdxl_turbo_4iter}
\end{figure*}

\begin{figure*}[p]
  \centering
  \setlength{\tabcolsep}{1pt}
  \renewcommand{\arraystretch}{0.88}
  \makebox[\textwidth][c]{%
  \begin{tabular}{@{}cccc@{}}
    \appvisimg{appendix/visualization/4iter/dmd2/iid/1.jpg}
    & \appvisimg{appendix/visualization/4iter/dmd2/iid/2.jpg}
    & \appvisimg{appendix/visualization/4iter/dmd2/iid/3.jpg}
    & \appvisimg{appendix/visualization/4iter/dmd2/iid/4.jpg} \\[-2pt]
    \appvisimg{appendix/visualization/4iter/dmd2/iid/5.jpg}
    & \appvisimg{appendix/visualization/4iter/dmd2/iid/6.jpg}
    & \appvisimg{appendix/visualization/4iter/dmd2/iid/7.jpg}
    & \appvisimg{appendix/visualization/4iter/dmd2/iid/8.jpg} \\[-2pt]
    \appvisimg{appendix/visualization/4iter/dmd2/iid/9.jpg}
    & \appvisimg{appendix/visualization/4iter/dmd2/iid/10.jpg}
    & \appvisimg{appendix/visualization/4iter/dmd2/iid/11.jpg}
    & \appvisimg{appendix/visualization/4iter/dmd2/iid/12.jpg} \\[-2pt]
    \appvisimg{appendix/visualization/4iter/dmd2/iid/13.jpg}
    & \appvisimg{appendix/visualization/4iter/dmd2/iid/14.jpg}
    & \appvisimg{appendix/visualization/4iter/dmd2/iid/15.jpg}
    & \appvisimg{appendix/visualization/4iter/dmd2/iid/16.jpg} \\
    \multicolumn{4}{c}{i.i.d.\ sampling} \\[4pt]
    \appvisimg{appendix/visualization/4iter/dmd2/mono/1.jpg}
    & \appvisimg{appendix/visualization/4iter/dmd2/mono/2.jpg}
    & \appvisimg{appendix/visualization/4iter/dmd2/mono/3.jpg}
    & \appvisimg{appendix/visualization/4iter/dmd2/mono/4.jpg} \\[-2pt]
    \appvisimg{appendix/visualization/4iter/dmd2/mono/5.jpg}
    & \appvisimg{appendix/visualization/4iter/dmd2/mono/6.jpg}
    & \appvisimg{appendix/visualization/4iter/dmd2/mono/7.jpg}
    & \appvisimg{appendix/visualization/4iter/dmd2/mono/8.jpg} \\[-2pt]
    \appvisimg{appendix/visualization/4iter/dmd2/mono/9.jpg}
    & \appvisimg{appendix/visualization/4iter/dmd2/mono/10.jpg}
    & \appvisimg{appendix/visualization/4iter/dmd2/mono/11.jpg}
    & \appvisimg{appendix/visualization/4iter/dmd2/mono/12.jpg} \\[-2pt]
    \appvisimg{appendix/visualization/4iter/dmd2/mono/13.jpg}
    & \appvisimg{appendix/visualization/4iter/dmd2/mono/14.jpg}
    & \appvisimg{appendix/visualization/4iter/dmd2/mono/15.jpg}
    & \appvisimg{appendix/visualization/4iter/dmd2/mono/16.jpg} \\
    \multicolumn{4}{c}{MoNO sampling}
  \end{tabular}}
  \caption{
  Qualitative comparison between i.i.d.\ sampling and MoNO (4-iter) using DMD2 for the prompt \textit{``a photo of a cup''}.
  }
  \label{fig:app_vis_dmd2_4iter}
\end{figure*}

\end{document}